\documentclass[journal,twoside,web]{ieeecolor}
\pdfoutput=1
\usepackage{tmi}
\usepackage{cite}
\usepackage{amsmath,amssymb,amsfonts}
\usepackage{graphicx}
\usepackage{textcomp}
\usepackage{hyperref}
\usepackage{booktabs}
\usepackage{bm}
\usepackage{capt-of}
\usepackage{mathtools}
\usepackage{siunitx}
\usepackage[figurename=Figure]{caption}
\usepackage{scrextend}
\usepackage{xspace}
\usepackage[table]{xcolor}
\usepackage{multirow}
\usepackage{wrapfig}

\usepackage{amsthm}
\usepackage{thmtools}
\usepackage{algorithm}
\usepackage{algpseudocode}
\usepackage[export]{adjustbox}
\usepackage{url}
\usepackage{subfig}
\usepackage{lineno}
\usepackage{pifont}
\usepackage{cite}
\usepackage[normalem]{ulem}
\usepackage{epstopdf}
\newcounter{algsubstate}
\renewcommand{\thealgsubstate}{\alph{algsubstate}}

\DeclareMathOperator*{\argmax}{arg\,max}
\DeclareMathOperator*{\argmin}{arg\,min}

\def\REVISION#1{\textcolor[RGB]{0,0,0}{#1}}

\newtheorem*{theorem*}{Theorem}

\renewcommand{\eqref}[1]{\mbox{Eq.~(\ref{#1})}}

\makeatletter
\DeclareRobustCommand\onedot{\futurelet\@let@token\@onedot}
\def\@onedot{\ifx\@let@token.\else.\null\fi\xspace}
\def\eg{\emph{e.g}\onedot}
\def\ie{\emph{i.e}\onedot}

\makeatother

\newcolumntype{L}[1]{>{\raggedright\arraybackslash}p{#1}}
\newcolumntype{C}[1]{>{\centering\arraybackslash}p{#1}}
\newcolumntype{R}[1]{>{\raggedleft\arraybackslash}p{#1}}

\newcommand{\cmark}{\ding{51}}%
\newcommand{\xmark}{\ding{55}}%

\renewcommand{\baselinestretch}{0.98}
\def\BibTeX{{\rm B\kern-.05em{\sc i\kern-.025em b}\kern-.08em
    T\kern-.1667em\lower.7ex\hbox{E}\kern-.125emX}}
\begin{document}

\title{Harvard Glaucoma Fairness: A Retinal Nerve Disease Dataset for Fairness Learning and Fair Identity Normalization}

\author{Yan Luo*, \IEEEmembership{Member, IEEE}, Yu Tian*, Min Shi*,  Louis R. Pasquale, Lucy Q. Shen, Nazlee Zebardast, \\Tobias Elze, and Mengyu Wang
\thanks{Yan Luo, Yu Tian, Min Shi, Tobias Elze, and Mengyu Wang are with Harvard Ophthalmology AI Lab, Schepens Eye Research Institute of Massachusetts Eye and Ear, Harvard Medical School, Boston, MA, USA. E-Mail: \{yluo16, ytian11, mshi6, tobias\_elze, mengyu\_wang\}@meei.harvard.edu.}
\thanks{Lucy Q. Shen and Nazlee Zebardast are with Massachusetts Eye and Ear, Harvard Medical School, Boston, MA, USA. E-Mail: \{lucy\_shen, nazlee\_zebardast\}@meei.harvard.edu.}
\thanks{Louis R. Pasquale is with Eye and Vision Research Institute, Icahn School of Medicine at Mount Sinai, New York, NY, USA. E-Mail: louis.pasquale@mssm.edu.}
\thanks{Yan Luo, Yu Tian, and Min Shi contributed equally as co-first authors.}
\thanks{Mengyu Wang is the corresponding author.}
% \thanks{T. C. Author is with the Electrical Engineering Department,
% University of Colorado, Boulder, CO 80309 USA, on leave from the National
% Research Institute for Metals, Tsukuba, Japan (e-mail: author@nrim.go.jp).}
}

\maketitle

\begin{abstract}

Fairness (also known as equity interchangeably) in machine learning is important for societal well-being, but limited public datasets hinder its progress. Currently, no dedicated public medical datasets with imaging data for fairness learning are available, though minority groups suffer from more health issues. To address this gap, we introduce Harvard Glaucoma Fairness (Harvard-GF), a retinal nerve disease dataset including 3,300 subjects with both 2D and 3D imaging data and balanced racial groups for glaucoma detection. Glaucoma is the leading cause of irreversible blindness globally with Blacks having doubled glaucoma prevalence than other races. We also propose a fair identity normalization (FIN) approach to equalize the feature importance between different identity groups. Our FIN approach is compared with various state-of-the-art fairness learning methods with superior performance in the racial, gender, and ethnicity fairness tasks with 2D and 3D imaging data, demonstrating the utilities of our dataset Harvard-GF for fairness learning. To facilitate fairness comparisons between different models, we propose an equity-scaled performance measure, which can be flexibly used to compare all kinds of performance metrics in the context of fairness.
The dataset and code are publicly accessible via \url{https://ophai.hms.harvard.edu/datasets/harvard-gf3300/}

\end{abstract}


%%%%%%%%% BODY TEXT
\section{Introduction}

\IEEEPARstart{M}{achine} learning research relies heavily on open-access datasets like ImageNet and CIFAR for its advancement \cite{deng2009imagenet,krizhevsky2009learning}. To conduct specific research in machine learning, it is essential to have access to datasets that are tailored to the particular research question. The topic of fairness (also known as equity interchangeably) in machine learning has garnered increasing attention in recent times, as it holds significant relevance to our society and human lives \cite{kadambi2021achieving,parikh2019addressing,Mehrabi_CSUR_2021}. In the field of fairness learning, the quantity and quality of publicly available datasets are limited.

\begin{figure}
  \centering
    \includegraphics[width=0.5\textwidth]{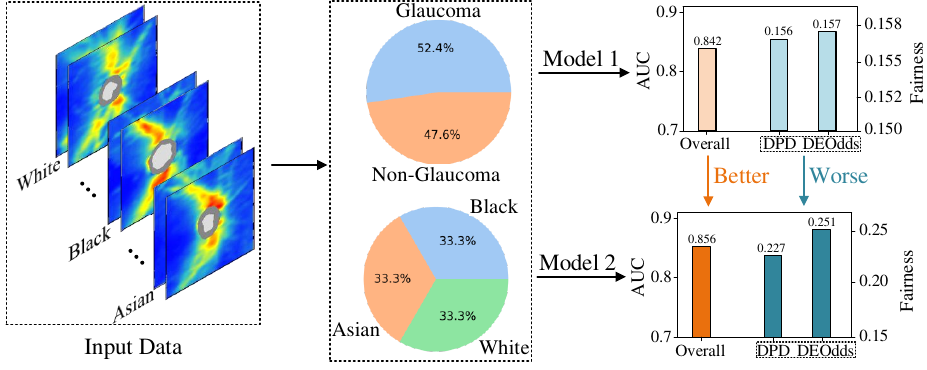}
    \vspace{-0.4cm}
  \caption{Illustration highlighting that fairness metrics such as DPD and DEOdds may not adequately account for the trade-off between accuracy and equity, even when the social identities associated with the samples are balanced. This misalignment is particularly problematic in safety-critical medical applications, which demand high accuracy. } 
   % \vspace{-0.7cm}
  \label{fig:problem}
\end{figure}

%Datasets that are publicly available in literature for studying fairness in machine learning can be categorized into five groups: criminology \cite{dressel2018accuracy, asuncion2007uci}, education \cite{wightman1998lsac,miao2010did,kuzilek2017open,asuncion2007uci}, finance \cite{asuncion2007uci,ruggles2015ipums, asuncion2007uci}, face recognition \cite{zhang2017age}, and healthcare \cite{asuncion2007uci}. Representative datasets include CAMPAS with 18,610 samples having identity attributes of gender, race, ethnicity and language in criminology \cite{dressel2018accuracy}, LASC with 20,798 samples having identity attributes of gender and race in education \cite{wightman1998lsac}, Adult with 48,842 samples having identity attributes of age and gender and KDD Census-Income with 299,285 samples having identity attributes of gender and race in finance \cite{asuncion2007uci}, UTKFace with 23,708 samples having identity attributes of age, gender and race in face recognition \cite{zhang2017age}, and Diabetes with 101,766 samples with an identity attribute of gender in healthcare \cite{asuncion2007uci}.

First, to date, only a handful of public fairness datasets have been utilized in at least three publications as shown in Table \ref{tbl:dataset} \cite{dressel2018accuracy, asuncion2007uci,wightman1998lsac,miao2010did,kuzilek2017open,ruggles2015ipums,zhang2017age}. Most of these datasets comprise tabular data, rendering them unsuitable for creating fair computer vision models requiring imaging data. This is particularly unsatisfactory given the prevalence of influential deep-learning models that rely on imaging data. The only imaging dataset proposed for the fairness learning problem is the UTKFace dataset \cite{zhang2017age}, which includes age, gender, and race information. Second, there are even fewer public fairness datasets in the area of healthcare and medical science. Two survey papers on fairness examined approximately 15 widely used public fairness datasets, and only two of them, the Heritage Health and Diabetes dataset \cite{asuncion2007uci}, are related to medical data and contain only tabular information. This is an inadequate representation of medical data available for fairness learning, given that minority groups have been reported to suffer more from health issues \cite{braveman2006health,marmot2008closing,marmot2007achieving,lyles2021focusing,carnethon2020disparities,marmot2012building,wong2015achieving}. It is crucial to investigate the disparities in deep learning prediction accuracy among different racial groups and take measures to minimize modeling bias if any. The third point is that the vast majority of datasets lack equal representation across different racial groups. This can make it challenging to determine the root cause of model performance discrepancy between different racial groups, as it may be unclear if the discrepancy stems from data imbalance or modeling bias.

In this paper, we introduce a new medical dataset called Harvard Glaucoma Fairness (Harvard-GF), which is a retinal nerve fiber layer (RNFL) dataset for glaucoma facilitating fairness learning. Note that, the modifier word ``Harvard" only indicates that our dataset is from the Department of Ophthalmology of Harvard Medical School and does not imply an endorsement, sponsorship, or assumption of responsibility by either Harvard University or Harvard Medical School as a legal identity. Glaucoma is an eye disease particularly suitable for fairness learning. Glaucoma is the number one cause of irreversible blindness globally \cite{tham2014global, quigley1996number, quigley2006number, tian2023fairseg, luo2023eye,shi2023jbhi,  luo2023harvard,shi2023artifact,sun2022time,Vermeer_TMI_2006,Joshi_TMI_2011,Cheng_TMI_2013,Fu_TMI_2017,Fu_TMI_2018,DiazPinto_TMI_2019}. The global prevalence of glaucoma for populations between 40 and 80 years old is 3.54\% affecting 80 million people \cite{tham2014global}. Racial minorities are affected disproportionately with Blacks having doubled glaucoma prevalence than other races. 
Unfortunately, early-stage glaucoma often does not present with any noticeable symptoms of vision loss, making professional vision tests crucial but inconvenient for patients to access as they require a visit to an ophthalmologist. Since glaucoma produces irreversible vision loss, early detection is critical to preserving patients' vision. Therefore, automatically screening glaucoma with retinal imaging using deep learning, which can be easily tested in pharmacies, is highly desirable. Before such a deep learning screening system can be used in practice, it has to be evaluated against modeling bias, which needs to be mitigated, if such biases exist.

% \yu{Our EyeFair dataset is the first fairness dataset in medical imaging with three racial (i.e., White, Black, and Asian) and two gender identities (i.e., Male and Female) for fairness study. We observe that deep learning systems also produce significantly inferior performance for the black racial group than the other two groups, even with balanced training data distribution. Furthermore, our dataset also explores the first 3D fairness benchmark to encourage a more comprehensive and clinically applicable fairness study.}

The highlights of our dataset are as follows: (1) The first fairness dataset for deep learning study in medical imaging; (2) The dataset has equal numbers of subjects from the three major racial groups of White, Black, and Asian, which avoids the data imbalance problem that may confound the fairness learning problem; (3) We have access to 3D imaging data in addition to 2D imaging data of RNFL thickness (RNFLT) maps derived from OCT. This provides the opportunity for 3D fairness learning, which has been a relatively unexplored area of study thus far.

In addition to our valuable dataset, we propose a fair identity normalization approach as an add-on contribution. Our fair identity normalization approach normalizes the logit features by each identity group such as racial and gender groups with learnable mean and standard deviation. This fair identity normalization approach aims to equalize the feature importance between different identity groups in deep learning modeling. Our fair identity normalization approach is compared with various state-of-the-art (SOTA) fairness learning methods in the literature. These evaluations of SOTA fairness learning methods and our proposed methods are expected to demonstrate the utilities of our dataset Harvard-GF for fairness learning. To facilitate fairness comparisons between different models, we propose an equity-scaled performance measure. The motivation to propose the equity-scaled performance measure is that current fairness metrics such as DPD and DEOdds may not adequately account for the trade-off between accuracy and equity as shown in Figure \ref{fig:problem}. In other words, a model with equally low accuracy for all identity groups could have high fairness, which is not reflected by DPD and DEOdds. Our equity-scaled performance measure can be flexibly used to compare all kinds of performance metrics in the context of fairness such as the area under the receiver operating characteristic curve (AUC) and accuracy.

Our core contributions are summarized as follows:
\begin{itemize}
\item We introduce the first dedicated fairness dataset with 2D and 3D medical imaging data.
\item We develop a novel fair identity normalization approach to equalize feature importance between different identity groups to improve model fairness.
\item We design a new equity-scaled metric to evaluate model performance penalized by fairness levels.
\end{itemize}

% \yu{
% To summarise, our contributions are the following: 
% \begin{itemize}
% \item We are the first and only real-world fairness dataset for deep learning study in medical imaging. 
% \item The dataset has equal numbers of subjects from the three major racial groups of White, Black, and Asian, which avoids the data imbalance problem that may confound the fairness learning problem.  
% \item In addition to the 2D imaging data, we introduce the first 3D imaging dataset and benchmark in fairness learning. 
% This provides the opportunity for 3D fairness learning, which has been a relatively unexplored area of study thus far.
% \item We propose a novel fair identity normalization to normalize the logit features for each identity group with customized learnable mean and standard deviation. 
% \item We propose a new equity-scaled performance measure to evaluate different fairness methods, enabling a more fair and applicable evaluation metric for medical imaging.  
% % Such an approach can be adapted to any off-the-shelf networks, data, and tasks, to enable a equalized feature space between different 
% \end{itemize}
% }

% The contributions can be summarized as follows. (1)  (2) The dataset has equal numbers of subjects from the three major racial groups of White, Black, and Asian, which avoids the data imbalance problem that may confound the fairness learning problem; (3) In addition to 2D imaging data, we have access to 3D imaging data, which is an unexplored area in fairness learning due to lack of data.
\section{Related Work}
\label{sec:related}

\noindent\textbf{Fairness Datasets}. Datasets that are publicly available in literature for studying fairness in machine learning can be categorized into five groups: criminology \cite{dressel2018accuracy, asuncion2007uci}, education \cite{wightman1998lsac,miao2010did,kuzilek2017open,asuncion2007uci}, finance \cite{asuncion2007uci,ruggles2015ipums}, face recognition \cite{zhang2017age}, and healthcare \cite{asuncion2007uci}. Representative datasets include CAMPAS with 18,610 samples having identity attributes of gender, race, ethnicity and language in criminology \cite{dressel2018accuracy}, LASC with 20,798 samples having identity attributes of gender and race in education \cite{wightman1998lsac}, Adult with 48,842 samples having identity attributes of age and gender and KDD Census-Income with 299,285 samples having identity attributes of gender and race in finance \cite{asuncion2007uci}, UTKFace with 23,708 samples having identity attributes of age, gender and race in face recognition \cite{zhang2017age}, and Diabetes with 101,766 samples with an identity attribute of gender in healthcare \cite{asuncion2007uci}.

\begin{table}[!t]
	\centering
	\caption{\label{tbl:dataset}
	   Public Fairness Datasets Commonly Used.
	}
  % \vspace{-0.3cm}
	\adjustbox{width=1\columnwidth}{
	\begin{tabular}{ L{10ex} C{24ex} L{24ex} C{10ex} C{8ex} C{4ex}}
		\toprule
		\textbf{Domain} & \textbf{Dataset} &   \textbf{Identity Attribute} &\textbf{Sample Size} & \textbf{Imaging Data} & \textbf{3D} \\

\cmidrule(lr){1-1} \cmidrule(lr){2-2} \cmidrule(lr){3-3} \cmidrule(lr){4-4}\cmidrule(lr){5-5}\cmidrule(lr){6-6}

        Criminology &  CAMPAS~\cite{dressel2018accuracy} & Gender; Race; Ethnicity; Language & 60,843 &\xmark  & \xmark \\
              Criminology  &  Communities and Crime~\cite{asuncion2007uci} & Race & 1,994 &\xmark  & \xmark \\
                                \cmidrule(lr){1-6}
        Education &  LASC~\cite{wightman1998lsac} & Gender; Race & 20,798 & \xmark  & \xmark \\
      Education &  Ricci~\cite{miao2010did} & Race & 118 & \xmark  & \xmark \\
        
         Education &  Student Performance~\cite{asuncion2007uci} & Age; Gender & 649 & \xmark  & \xmark \\
                  Education &  OUTLAD~\cite{kuzilek2017open} & Gender & 32,593 & \xmark  & \xmark \\
                  \cmidrule(lr){1-6}
           Finance &  Adult~\cite{asuncion2007uci} & Age; Gender & 48,842 & \xmark  & \xmark \\
            Finance & Bank Marketing~\cite{asuncion2007uci} & Age; Marital & 45,211 &\xmark  & \xmark \\
            Finance & Credit Card Clients~\cite{asuncion2007uci} & Gender; Marriage; Education & 30,000 & \xmark & \xmark \\
            Finance & Dutch Census~\cite{ruggles2015ipums}  & Gender & 60,420 & \xmark & \xmark \\
             Finance & German Credit~\cite{asuncion2007uci}  & Age; Gender & 1,000 & \xmark & \xmark \\
            Finance &  KDD Census-Income~\cite{asuncion2007uci} & Gender; Race & 299,285 & \xmark & \xmark \\
         \cmidrule(lr){1-6}
            Face Recognition &  UTKFace~\cite{zhang2017age} & Age; Gender; Race & 23,708 & \cmark & \xmark \\
         \cmidrule(lr){1-6}
         Healthcare & Diabetes~\cite{asuncion2007uci}  & Gender & 101,766 & \xmark & \xmark \\
        Healthcare &  Heritage Health ~\cite{asuncion2007uci} & Age & 147,473 & \xmark & \xmark \\
          \midrule
    Healthcare & Harvard-GF & Age; Gender; Race; Ethnicity; Language; Marriage & 3,300 & \cmark & \cmark \\
		\bottomrule	
	\end{tabular}}
\end{table}

Table \ref{tbl:dataset} highlights that fairness learning has mainly focused on tabular data in prior datasets except for the UTKFace dataset \cite{zhang2017age}. However, only having tabular data limits the fairness evaluation and improvement of the vast majority of deep learning models that rely on imaging inputs. This is particularly concerning since imaging data are essential in developing deep learning models. In addition, it is disappointing to note that public fairness datasets in healthcare are limited and typically lack imaging data. This is especially concerning given that medical imaging data are pivotal in current disease diagnosis practices. Furthermore, it is worth noting that the previous datasets were not originally intended to explore fairness and were repurposed for this purpose. As a result, these datasets frequently have an uneven distribution of subjects from different racial groups, which can complicate the fairness learning problem with the data imbalance issue \cite{yu2022re,cui2019class,cao2019learning,dong2018imbalanced}. Lastly, prior datasets typically have very few identity attributes for fairness learning purposes. For instance, the LASC dataset only includes gender and race as identity attributes, while the Diabetes dataset only contains gender as a single identity attribute. Such limited identity attributes constrain the datasets' capacity to address various fairness issues related to identity traits. In comparison, our new dataset Harvard-GF will address these unmet demands with our 2D and 3D imaging data and evenly distributed racial groups, which is elaborated in Section \ref{sec:related}.

% [put a table of datasets here]

% \cite{Mehrabi_CSUR_2021,Shui_NeurIPS_2022}

% Expected fairness violations \cite{Zietlow_CVPR_2022}

\noindent\textbf{Fairness Models}. Computer vision datasets and methods often produce biased predictions due to data/sampling inequalities~\cite{Mehrabi_CSUR_2021}, which has drawn significant attention in the community. Such unfairness can be generally alleviated by introducing strategies through three different training stages, including pre-processing (i.e. data de-bias), in-processing (model de-bias), and post-processing (prediction de-bias). 
% This paper focuses on the model de-bias category most relevant to our work.
Pre-processing methods~\cite{quadrianto2019discovering,ramaswamy2021fair,zhang2020towards,park2022fair} focus on ``de-biasing" the training data before training begins, with the expectation that balanced and fair training sets could potentially produce fair models. Existing methods rely on either data representation transformation to enforce the model to discard the feature representation correlations~\cite{quadrianto2019discovering,park2022fair} between different sensitive attributes 
% through adversarial learning~\cite{}, disentangled learning, and contrastive learning~\cite{}, 
or data distribution augmentation~\cite{ramaswamy2021fair,zhang2020towards,Zietlow_CVPR_2022} that generates data samples to balance the training distribution over different sensitive attributes (i.e., races or genders). However, those pre-processing methods require an inefficient two-stage training/processing that can potentially jeopardize the computational efficiency, and recent SOTA contrastive learning approaches~\cite{park2022fair} rely on data augmentations (i.e., color permutations, color distortions, sobel filtering, etc.) are difficult to be adapted in different medical domains (i.e., 3D CT data). 
In-processing methods aim to achieve fairness during the training procedure by introducing fairness constraints to penalize models' ability to distinguish sensitive attributes~\cite{beutel2017data,roh2020fr,sarhan2020fairness,zafar2017fairness,zhang2018mitigating}. Specifically, authors in \cite{zhang2018mitigating,beutel2017data,roh2020fr} propose to use adversarial training to deteriorate the model ability to distinguish the protected attributes. Sarhan et al.~\cite{sarhan2020fairness} enforce disentanglement constraints on low-dimensional feature representations to make the representations of different attributes indistinguishable. Zafar et al.~\cite{zafar2017fairness} set constraints to enforce the model to achieve good accuracy that also incurs the least possible disparate impact. 
Nevertheless, those methods that explicitly manipulate the loss functions during training can sacrifice the overall accuracy of target labels, and those applied constraints/penalties are often task-specific, which can be in-adaptable to different tasks (e.g., Glaucoma detection and progression forecasting) and high-dimensional data domains (e.g., high-dimensional 2D and 3D medical images). 
Post-processing methods tend to adjust the model's predictions to achieve fairness after the training process~\cite{wang2022fairness,kim2019multiaccuracy}. However, those methods manipulate sensitive attributes, trained models, or data during testing time, which hinders applicability in various computer vision and medical imaging tasks. 
On the contrary, we propose a simple, effective, and plug-and-play normalization module to obtain fairness during training that is adaptable to different types of tasks and medical data. 

\noindent\textbf{Fairness Metrics}. Fairness is a multi-dimensional concept that can be defined and interpreted in different ways depending on a particular application's specific context and goals. The most common form of fairness is group fairness, which means impartially treating people from the same biological or social groups. However, optimizing for group fairness may come at the cost of reduced individual fairness evidenced by reduced prediction accuracy.

There are three common fairness metrics that are based on different assumptions. They are demographic parity difference (DPD) \cite{Bickel_Science_1975,Agarwal_ICML_2018,Agarwal_ICML_2019}, difference in equal opportunity (DEO) \cite{Hardt_NeurIPS_2016}, and difference in equalized Odds (DEOdds) \cite{Agarwal_ICML_2018}. Demographic parity \cite{Agarwal_ICML_2018,Agarwal_ICML_2019} is designed to guarantee that a predictive model's predictions are not influenced by an individual's membership in a sensitive group. This means that demographic parity is accomplished when there is no correlation between the probability of a specific prediction and an individual's sensitive group membership. Correspondingly, a DPD of 0 indicates that all groups have the same selection rate. DEO \cite{Hardt_NeurIPS_2016} focuses on the true positive rate (TPR) of the predictive model for different groups defined by a sensitive attribute (such as race or gender). Equal opportunity is achieved when the TPR is the same across all groups, meaning that the classifier makes positive predictions at the same rate for members of each group who truly belong to the positive class. DEOdds \cite{Agarwal_ICML_2018} is a generalized version of DEO. It requires that the predictive model’s predictions are independent of sensitive group membership with sensitive groups having the same false positive rates and true positive rates.

This paper focuses on safety-critical medical scenarios where a significant compromise of reduced model accuracy harming individual fairness in a tradeoff for improved group fairness is unacceptable. More specifically, an improvement in these fairness metrics of DPD, DEO, and DEOdds may significantly lower AUC or accuracy, which is unacceptable for individual patients. To address this issue, we propose an equity scaling mechanism to describe model accuracy scaled for group fairness.

\noindent\textbf{Ethical Implications of Fairness Techniques in Medical Artificial Intelligence (AI)}.
\REVISION{Medical AI models have enabled affordable large-scaled disease screenings\cite{gargeya2017automated, gulshan2016development, thompson2020review}, which can greatly benefit the minority and socioeconomically disadvantaged groups that often do not have access to expensive quality healthcare resources to reduce health disparities. Model fairness needs to be optimized before practical deployment to maximize the benefit of medical AI for reducing health disparities. Optimizing model fairness is a complex issue in disease screening with medical AI, as it is undesirable to improve the model performance in one demographic group at the cost of decreasing model performance in other demographic groups which poses big ethical concerns. Fairness techniques shall be designed in a way that at least does not decrease overall and group-specific model performance, yet there may be a tradeoff between overall and group-specific model performance and group performance disparity. Therefore, we propose the equity-scaled metrics to account for this potential tradeoff to fairly compare different fairness learning techniques.
%The ethical implications of fairness techniques in medical AI are multifaceted, requiring careful consideration to protect patient privacy, ensure informed consent, and address potential impacts on diverse populations. Privacy concerns necessitate robust data security and anonymization measures, while transparent communication and opt-in/opt-out mechanisms are crucial for obtaining informed consent from patients \cite{filkins2016privacy}. Mitigating biases in algorithms is central to fairness, demanding ongoing evaluation and monitoring, along with a commitment to equitable access and representation in data \cite{yang2023algorithmic}. Explainability and accountability are essential for building trust, and the establishment of ethics committees and regulatory frameworks can provide oversight and guidance \cite{MARKUS2021103655}. Ultimately, a comprehensive and collaborative approach, involving diverse stakeholders, is imperative to navigate the ethical complexities of integrating fairness techniques into medical AI responsibly.
}
% In this work, we focus on safety-critical medical scenarios. A mainstream performance metric is Area under the ROC Curve (AUC) in medical applications.

% [Refer to \url{https://fairlearn.org/main/user_guide/assessment/common_fairness_metrics.html}]

% [Discuss why DEO and DDP are not competent to depict the trade-off between fairness and performance.]
\section{Dataset Analysis}
\label{section:data}
\subsection{Data Collection and Quality Control}
Our institute's institutional review board (IRB) approved this study, which followed the principles of the Declaration of Helsinki. Since the study was retrospective, the IRB waived the requirement for informed consent from patients. Note that, the modifier word ``Harvard" from the dataset name only indicates that our dataset is from the Department of Ophthalmology of Harvard Medical School and does not imply an endorsement, sponsorship, or assumption of responsibility by either Harvard University or Harvard Medical School as a legal identity.

% This study was approved by the Massachusetts Eye and Ear institutional review board and  
% adhered to the tenets set forth by the Declaration of Helsinki. Given the retrospective nature of  
% this study using existing clinical data, the need for informed consent was waived.

\begin{figure}
  \centering
    \includegraphics[width=0.5\textwidth]{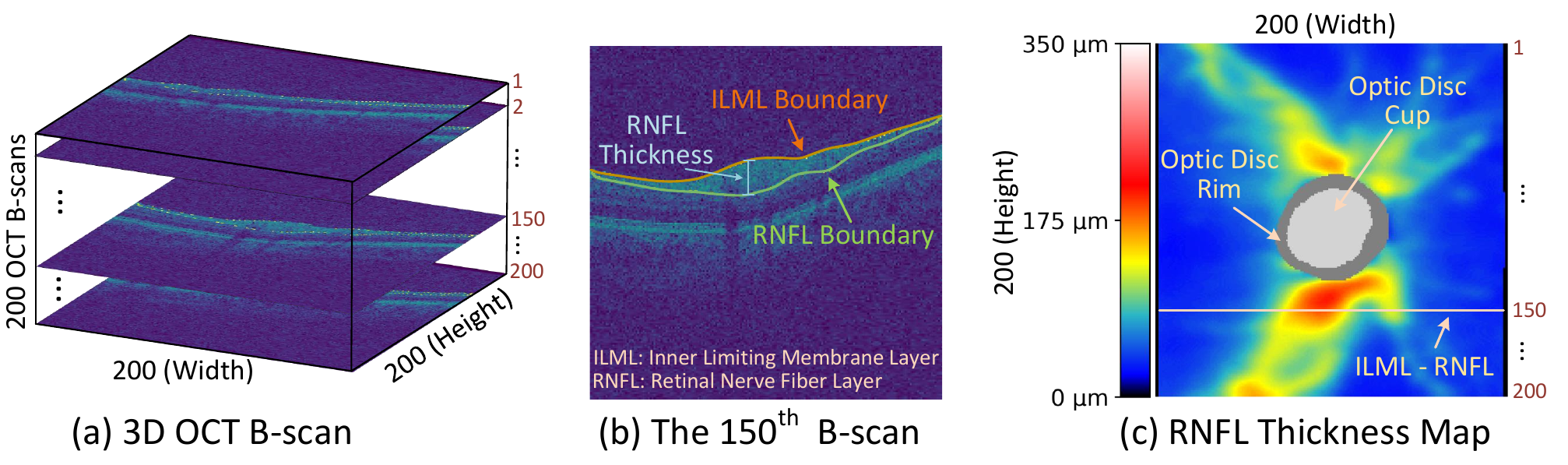}
    \vspace{-0.5cm}
  \caption{Illustrations that depict RNFLT maps, OCT B-scans images, and the relationship between the two data types.} 
   % \vspace{-0.7cm}
  \label{fig:oct_rnflt}
\end{figure}

The subjects tested between 2010 and 2021 are from a large academic eye hospital. There are three types of data to be released in this study: (1) optical coherence tomography (OCT) scans; (2) patient demographics; (3) glaucoma diagnosis defined based on visual field tests. RNFLT maps, OCT B-scans, and the relationship between the two types of data are illustrated in Figure \ref{fig:oct_rnflt}. We detail each type of data as follows. (1) OCT scans are state-of-the-art 3D imaging tests for diagnosing various eye diseases such as glaucoma, age-related macular degeneration, and diabetes retinopathy. For glaucoma, the 2D RNFLT map derived from the 3D OCT B-scans (Cirrus, Carl Zeiss Meditec, Jena, Germany) calculated as the vertical distance between the inner limiting membrane layer boundary and retinal nerve fiber layer boundary segmented by the manufacturer's software was the standard structural measurement used by clinicians. The 2D RNFLT map has a resolution of $200 \times 200$ pixels within an area of $6 \times 6$ mm$^2$ around the optic disc. The RNFLT range is between 0 and 350 microns. In addition to the 2D RNFLT maps, we also release the 3D OCT B-scans. To control data quality, OCT scans with a signal strength of less than 6 (10 represents the best imaging quality) were excluded. (2) Patient demographics released include age, gender, race, ethnicity, language proficiency, and marital status. This dataset is designed to be heavily focused on studying racial fairness. Therefore we prioritize to control that we have the same subject numbers for each racial group from the hospital data pool. (3) Glaucoma diagnosis is defined based on the 24-2 visual field test by Humphrey Field Analyzer (Carl Zeiss Meditec, Dublin, CA), which measures visual sensitivity spatially with a radius of 24 degrees from the fixation for each eye of a patient. Based on the manufacturer's guidelines, we only include reliable visual field tests with criteria commonly used in clinical practice.  This dataset can only be used for non-commercial research purposes. At no time, the dataset shall be used for clinical decisions or patient care. The data use license is CC BY-NC-ND 4.0.

\subsection{Data Characteristics}

Our Harvard-GF dataset contains 3,300 samples from 3,300 subjects with 1,748 subjects having glaucoma. We divide our data into the training set with 2,100 samples, the validation set with 300 samples, and the test set with 900 samples. \REVISION{For each subject, we always selected the last visit's data and randomly selected one eye per subject.}
% The racial groups (\ie Asian, Black, and White) are balanced for ruling out the negative effects caused by race imbalance. This gives rise to 1,812 female patients and 1,488 male patients, 3,025 non-hispanic patients and 87 hispanic patients. There are 188 patients with missed ethnicity so we exclude them from the training/evaluation process. Moreover, the average age of patients is over 60-year-old. More information can be found in Figure \ref{fig:data_charateristics}.

\begin{figure}[!t]
	\centering
	\subfloat[]{\includegraphics[width=0.19\textwidth]{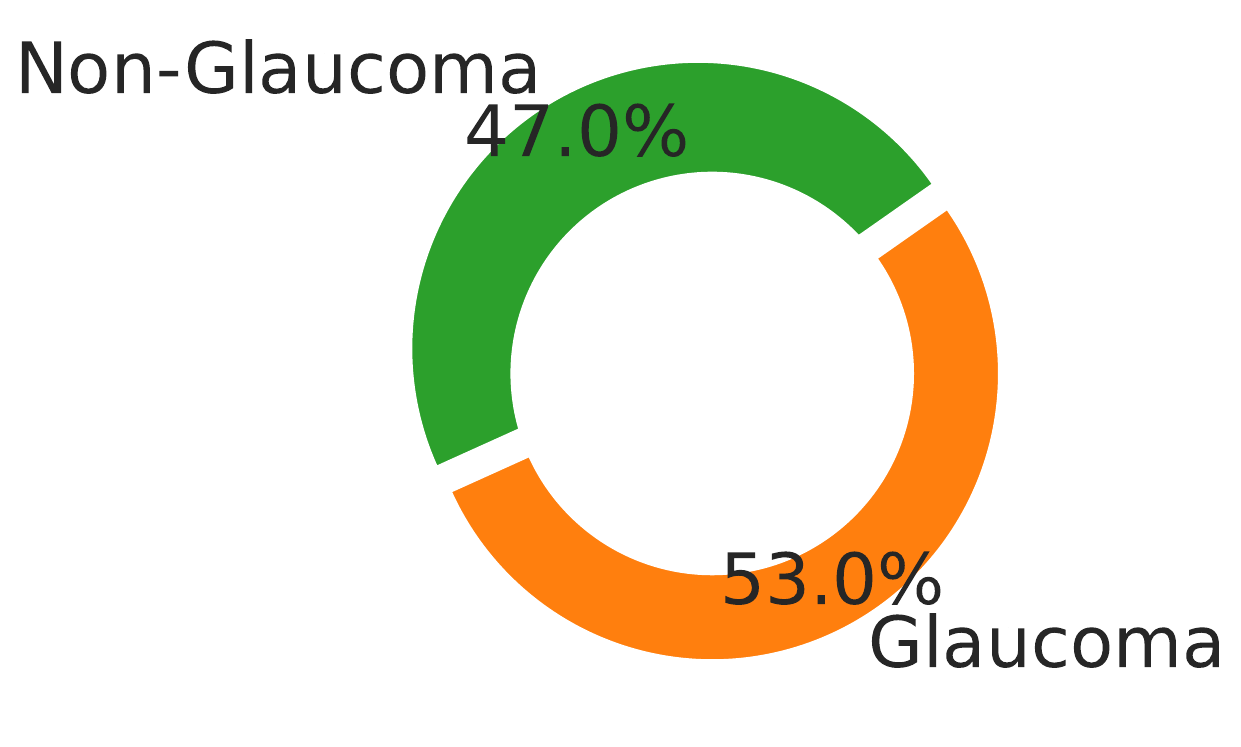}  \label{fig:dist_label}} \hfill
	\subfloat[]{\includegraphics[width=0.14\textwidth]{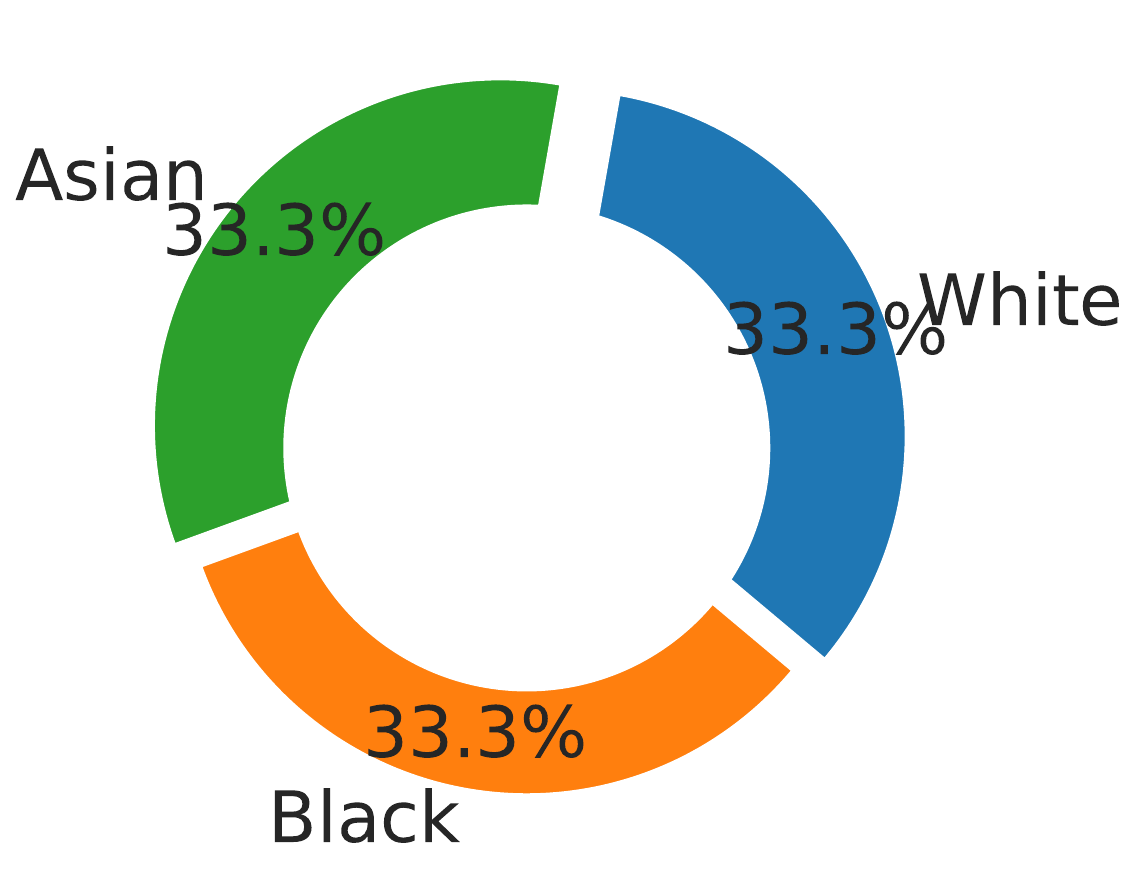}  \label{fig:dist_race}} \\
        \subfloat[]{\includegraphics[width=0.17\textwidth]{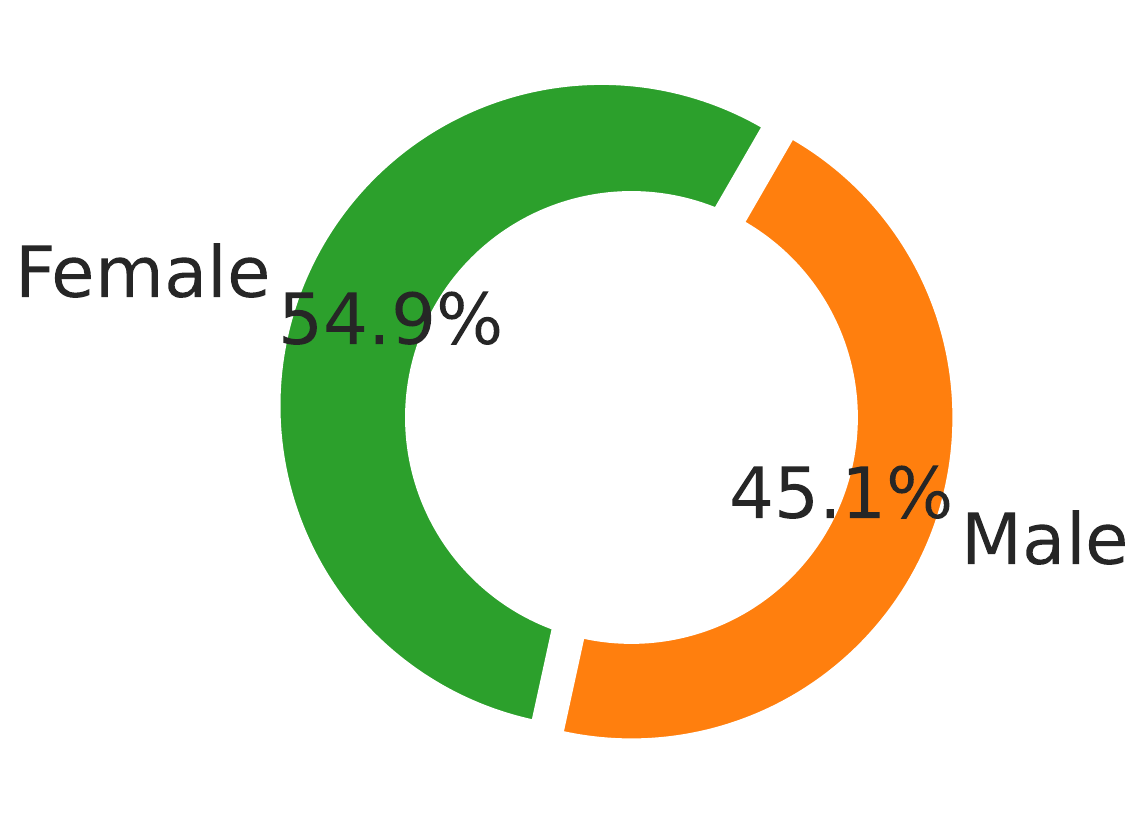} \label{fig:dist_gender}} \hfill
        \subfloat[]{\includegraphics[width=0.12\textwidth]{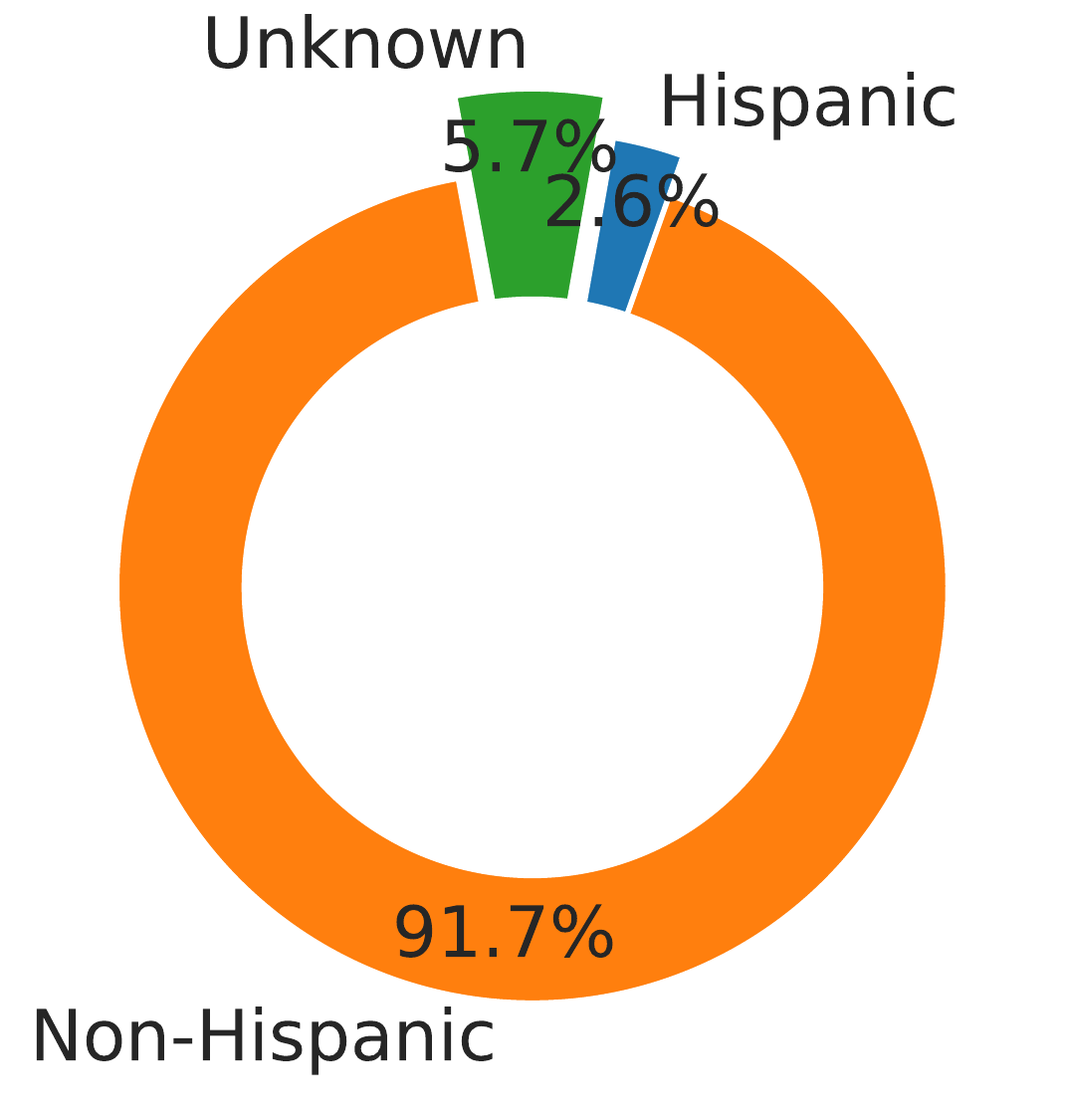}
        \label{fig:fig:dist_eth}} 
        \subfloat[]{\includegraphics[width=0.16\textwidth]{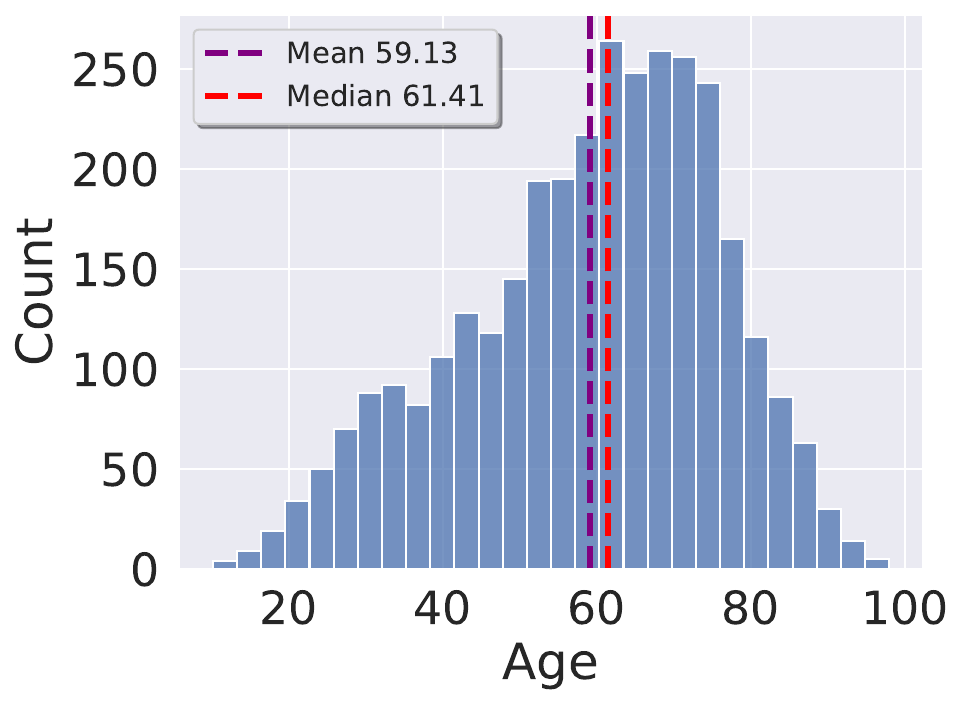}
        \label{fig:fig:dist_age}}
	\caption{\label{fig:data_charateristics}
    	The distributions of the samples categorized by various factors, including glaucoma class (a), race (b), gender (c), ethnicity (d), and age (e).
    	}
\end{figure}

\begin{figure}
  \centering
    \includegraphics[width=0.4\textwidth]{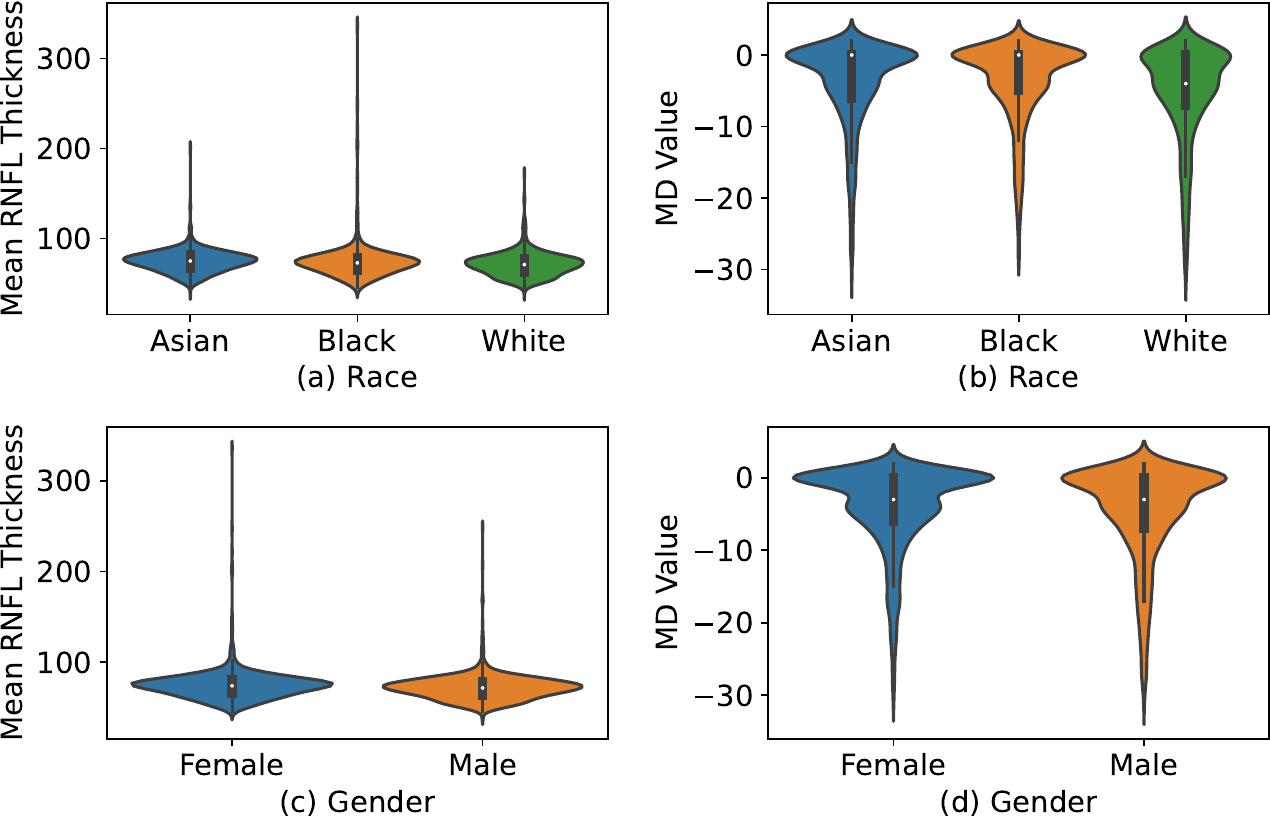}
    \vspace{-0.2cm}
  \caption{The distributions of retinal nerve fiber layer thickness and vision loss severities measured by mean deviation against different racial and gender groups.} 
   % \vspace{-0.7cm}
  \label{rnfltdist}
\end{figure}

The average ages, RNFLTs, and vision loss severities measured by the decibel unit are 58.7 $\pm$ 16.6 years, 85 $\pm$ 22 microns, and -4.6 $\pm$ 6.2 dB, respectively. Our dataset has three racial groups: Asia, Black, and White. Each racial group contains 1,000 samples. 54.9\% of the subjects are female. The distribution of ethnicity, language proficiency, and marital status are detailed as follows. Ethnicity: 91.5\% Non-Hispanic, 2.6\% Hispanic, and 5.7\% unknown. Language proficiency: 87.4\% English speakers, 0.01\% Spanish speakers, 10.7\% other language speakers, and 0.8\% unknown. Marital status: 56.7\% marriage/partnered, 28.8\% single, 5.4\% divorce, 1.2\% legally separated, 5.3\% widowed, and 2.6\% unknown. More details can be found in Figure \ref{fig:data_charateristics}.

In this paper, we focus on studying fairness learning with respect to race and gender. The prevalence of glaucoma in Asians, Blacks, and Whites are 47.4\%, 61.4\%, and 48.4\%, respectively. Blacks have significantly (p $<$ 0.001) higher glaucoma prevalence than Whites and Asians. The prevalence of glaucoma in females and males are 51.1\% and 54.0\%, respectively, which do not differ significantly (p = 0.12). \REVISION{As shown in Figure \ref{rnfltdist}, Blacks have thinner RNFLTs (70.9 $\pm$ 13.1 µm) and worse vision loss (i.e., least MD values) than Whites (73.7 $\pm$ 19.4 µm) and Asians (74.4 $\pm$ 11.8 µm) significantly (p $<$ 0.001 for both), while there are no significant differences between Whites and Asians. Regarding gender, males (71.6 $\pm$ 13.9 µm) have thinner (p $<$  0.001) RNFLTs and worse (p = 0.002) vision loss than females (74.2 $\pm$ 16.1 µm) significantly, though the glaucoma prevalence does not differ between males and females as described in the above paragraph.}

\section{Methodology for Understanding Debiasing}
\label{sec:method}

Ophthalmic imaging data from individuals of the same identity group often exhibit similar patterns. This is due to genetic and environmental factors that result in similar retinal traits and disease pathways. In order to address this identity-based similarity, we propose a fair identity normalization method to enhance features according to the statistical properties of the individual's identity, which would ultimately produce fairer prediction results.

In addition, conventional fairness metrics like the difference of demographic parity \cite{Bickel_Science_1975,Agarwal_ICML_2018,Agarwal_ICML_2019} and equalized odds \cite{Agarwal_ICML_2018} may not be suitable for medical applications for two reasons. Firstly, these metrics might not always align with other important metrics, such as accuracy and efficiency. A model optimized for fairness may sacrifice accuracy, which is unacceptable in safety-critical medical applications. Secondly, these metrics' effectiveness heavily relies on the accuracy and representativeness of the data used. Biased, incomplete, or inaccurate data may cause these metrics to fail in identifying and addressing unfairness.

% members of the same racial group may share experiences of discrimination, marginalization, and social inequality, which can impact their attitudes, behaviors, and social interactions.

% To comprehensively understand fairness in EyeFair in terms of demographic information and gender, we propose a fair-identity normalization method to 

In the subsequent subsections, we begin by defining fair identity normalization to examine the impacts of enhancing the features in an identity-dependent manner. Subsequently, we introduce an equity scaling mechanism that considers the influence of identity-dependent equity on prominent performance metrics, such as accuracy and AUC.

\subsection{Fair Identity Normalization}

We denote $x\in \mathcal{X} \subset \mathbb{R}^{d}$ as a patient sample, and $y\in \mathcal{Y} = \{0, 1\}$ as a glaucoma label. Similar to the settings in previous fairness learning methods \cite{quadrianto2019discovering,ramaswamy2021fair,Zietlow_CVPR_2022}, $a\in \mathcal{A}$ is the attribute associated with the sample. Specifically, the attribute set $\mathcal{A}$ in Harvard-GF can be types of social identities, \ie race ($\mathcal{A}=\{\text{Asian}, \text{Black or African American}, \text{White or Caucasian}\}$) or gender ($\mathcal{A}=\{\text{Male}, \text{Female}\}$). For simplicity, we digitize these identities, \ie $\mathcal{A}=\{0, 1, 2\}$ or $\mathcal{A}=\{0, 1\}$. Also, we have a model $f: \mathbb{R}^{d} \xrightarrow{\theta} \mathcal{Y}$ to predict the glaucoma label. 

\begin{figure}
  \centering
    \includegraphics[width=0.5\textwidth]{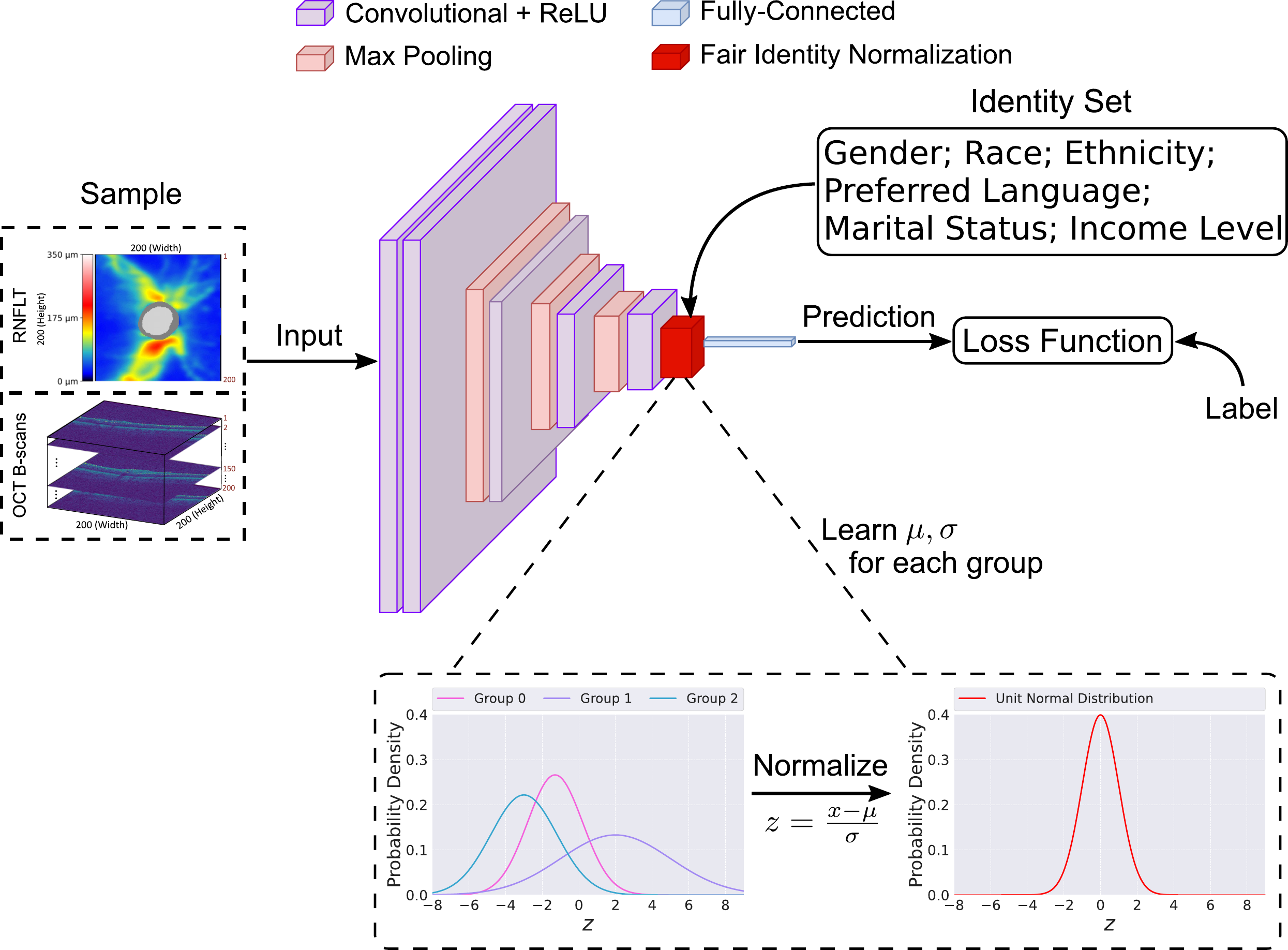}
    \vspace{-0.5cm}
  \caption{Schematic view of the proposed fair identity normalization.} 
   % \vspace{-0.7cm}
  \label{fig:schematic}
\end{figure}

We hypothesize that the samples associated with a certain identity have an underlying correlation with each other. Therefore, as shown in Figure \ref{fig:schematic}, we propose a fair identity normalization (FIN) method to enhance the discriminative features of the samples with the same identity. To this end, the model can be generally viewed as two components: the backbone $f_{\theta_{b}}(\cdot)$ that generates the discriminative features $z=f_{\theta_{b}}(x)$, and the final linear module $f_{\theta_{h}}(\cdot)$ that maps the features to the logits $z'=f_{\theta_{h}}(z)$. Instead of ignoring the identity $a$, the proposed FIN $\xi$ takes $z$ and $a$ as input to produce estimated statistics
\begin{equation}
    \begin{split}
    \hat{z} = \xi(z, a; \mu_{A}, \sigma_{A}) = \frac{z-\mu_{A}}{\sigma_{A}}, \ \ \ \ a=A \in \mathcal{A}
\end{split}
\label{eqn:es_z_norm}
\end{equation}
where $A$ represents a certain identity, \eg, Black, Female, etc, $\mu_{A}$ and $ \sigma_{A}$ are the mean and standard deviation with the same dimensionality as $z$. They are correlated to all the samples with identity $A$. Then, $z' = f_{\theta_{h}}(\hat{z})$. $\mu_{A}$ and $\sigma_{A}$ are learnable such that the proposed FIN is easily suited to the conventional end-to-end learning framework. This process can be mathematically formulated as
\begin{equation}
    \begin{split}
    \mu_{A} = \frac{\partial \ell}{\partial z'} \frac{\partial z'}{\partial \mu_{A}}, \hspace{2ex} \sigma_{A} = \frac{\partial \ell}{\partial z'} \frac{\partial z'}{\partial \sigma_{A}}, \hspace{2ex} \forall \{(x, a, y)|a=A\}
\end{split}
\label{eqn:fin_mu_mean}
\end{equation}
Without the restrictions on $\sigma_{A}$, $\sigma_{A}$ can be updated as a negative value along the learning process. Negative $\sigma_{A}$ is not valid in the context of the normalization with the standard deviation. Following \cite{Blundell_ICML_2015}, we apply a reparameterization trick to make $\sigma_{A}$ non-negative, \ie, $\sigma_{A} = log(1 + exp(\tau_{A}))$. Furthermore, for the sake of numerical stability, we follow the batch normalization \cite{Ioffe_ICML_2015} to use a momentum $m \in [0, 1]$ to integrate $z$ and $\hat{z}$, \ie, 
% $z' = (1-m)\cdot \hat{z} + m\cdot z$. 
\begin{equation}
    \begin{split}
    z' = (1-m)\cdot \hat{z} + m\cdot z
\end{split}
\label{eqn:momentum}
\end{equation}
We use a unit normal distribution to initialize $\mu_{A}, \sigma_{A}$. The updated logit $z'$ takes the characteristics of identity-wise features into account to adaptively enhance the predictions. \REVISION{The workflow of FIN is summarized in Algorithm \ref{algo:FIN}.}
% $\tau(z; \mu, \sigma)$

\REVISION{
The proposed FIN shows promising potential for generalization across medical imaging domains beyond glaucoma detection. At its core, FIN offers a flexible framework to normalize feature distributions on a demographic subgroup basis to mitigate accuracy disparities. This concept of identity-tailored normalization to enhance equity is broadly applicable to other visual diagnostic tasks, especially as medical specialties increasingly rely on advanced imaging data where algorithmic biases can emerge. As diagnostic AI expands, techniques like FIN that promote equitable model performance for all patient subgroups are crucial for safe, ethical deployment in the high-stakes healthcare setting. In summary, the versatility of FIN’s conceptual approach makes it well-suited for adaptation to improve fairness in a range of medical imaging applications.
}

\begin{algorithm}[t]
\caption{Fair Identity Normalization}
\begin{algorithmic}[1]
\State \textbf{Input:} Sample $x \in \mathcal{X}$, label $y \in \mathcal{Y}$, attribute $a \in \mathcal{A}$, model $f: \mathbb{R}^{d} \xrightarrow{\theta} \mathcal{Y}$
\State \textbf{Output:} Logit $z' = f_{\theta_{h}}(\hat{z})$

\State \textbf{Initialization:} $\mu_{A}$, $\sigma_{A}$ are initialized with unit normal distribution
\State $z = f_{\theta_{b}}(x)$ \Comment{Backbone component}
\State $\hat{z} = \xi(z, a; \mu_{A}, \sigma_{A})$ \Comment{Normalization}

\For{$\{(x, a, y)|a=A\}$}
    \State Update $\mu_{A} \gets \frac{\partial \ell}{\partial z'} \frac{\partial z'}{\partial \mu_{A}}$
    \State Update $\tau_{A} \gets \text{log}(1 + \text{exp}(\tau_{A}))$ \Comment{Reparameterization trick}
\EndFor

\State Update $z' \gets (1-m)\cdot \hat{z} + m\cdot z$ \Comment{Batch normalization with momentum $m$}

\State \textbf{Return:} $z'$
\end{algorithmic}
\label{algo:FIN}
\end{algorithm}

\begin{table*}[!t]
	\centering
	\caption{\label{tbl:rnflt_race}
	    Performance on the test set of RNFLT maps with \textbf{race} identities. All scores are in the form of percentages. Each experiment is run with 5 random seeds. 
     % The baseline method is the supervised model trained with labeled samples only, while pseudo-sup stands for the proposed \textit{pseudo supervisor} method.
	}
 % \vspace{-0.3cm}
	\adjustbox{width=2\columnwidth}{
	\begin{tabular}{L{15ex} C{12ex} C{12ex} C{12ex} C{12ex} C{12ex} C{12ex} C{12ex} C{12ex} C{12ex}}
		\toprule
         & \textbf{Overall} & \textbf{Overall} & \textbf{Overall} & \textbf{Overall} & \textbf{Asian} & \textbf{Black} & \textbf{White} & \textbf{Overall} & \textbf{Overall} \\
		\textbf{Method} & \textbf{ES-Acc$\uparrow$} & \textbf{Acc$\uparrow$} & \textbf{ES-AUC$\uparrow$} & \textbf{AUC$\uparrow$} & \textbf{AUC$\uparrow$} & \textbf{AUC$\uparrow$} & \textbf{AUC$\uparrow$} & \textbf{DPD$\downarrow$} & \textbf{DEOdds$\downarrow$} \\
		\cmidrule(lr){1-1} \cmidrule(lr){2-2} \cmidrule(lr){3-3} \cmidrule(lr){4-4} \cmidrule(lr){5-8} \cmidrule(lr){9-9} \cmidrule(lr){10-10}
        Adv~\cite{beutel2017data} & 68.46$\pm$1.59   & 76.22$\pm$1.78 & 77.86$\pm$1.19 & 84.67$\pm$0.88  &  88.08$\pm$0.94  & 80.01$\pm$1.28  & 84.55$\pm$0.91 &9.88$\pm$3.56 & 9.64$\pm$3.92 \\
        SimCLR~\cite{chen2020improved} & 68.54$\pm$0.78   & 73.37$\pm$3.01 & 79.76$\pm$0.64 & 86.08$\pm$0.13  &  87.05$\pm$1.74  & 79.08$\pm$0.39  & 84.23$\pm$1.12 & 8.88$\pm$1.39 & 9.33$\pm$2.82 \\
        SupCon~\cite{khosla2020supervised} & 68.46$\pm$2.13   & 75.92$\pm$1.26 & 77.92$\pm$0.63 & 84.74$\pm$0.72  &  88.18$\pm$1.53  & 79.61$\pm$0.69  & 84.94$\pm$0.98  & 9.56$\pm$2.87 & 9.64$\pm$3.92 \\ \midrule
        FSCL~\cite{park2022fair} & 71.04$\pm$0.73   & 76.92$\pm$0.59 & 77.99$\pm$0.51 & 84.61$\pm$0.89  &  87.54$\pm$1.32  & 79.63$\pm$0.21  & 85.17$\pm$0.31 & 14.33$\pm$2.59 & 13.41$\pm$4.87 \\
        FSCL+~\cite{park2022fair} & 69.58$\pm$0.78   & 75.85$\pm$1.32 & 78.04$\pm$0.87 & 83.96$\pm$1.17  &  87.05$\pm$1.74  & 79.75$\pm$0.39  & 84.24$\pm$1.29  & \textbf{9.11}$\pm$0.35 & 7.24$\pm$0.72 \\ 
        FSCL+ w/ FIN & \textbf{74.04}$\pm$0.67 & 77.02$\pm$0.81 & 78.69$\pm$0.55 & \textbf{86.40}$\pm$0.73 & \textbf{88.39}$\pm$1.13 & 81.17$\pm$0.43 & \textbf{88.96}$\pm$0.97 & 10.67$\pm$0.88 & \textbf{6.33}$\pm$1.12 \\
        \midrule
        No Norm & 69.40$\pm$1.21 & 77.08$\pm$0.56 & 77.36$\pm$0.93 & 84.97$\pm$0.58 & 88.00$\pm$0.76 & 79.46$\pm$0.84 & 86.28$\pm$0.85 & 11.99$\pm$2.01 & 13.93$\pm$2.45  \\
        BN \cite{Ioffe_ICML_2015} & 69.01$\pm$1.71 & 77.13$\pm$1.08 & 76.86$\pm$0.58 & 84.39$\pm$0.58 & 87.51$\pm$1.11 & 79.27$\pm$0.40 & 84.93$\pm$2.07 & 12.46$\pm$3.07 & 14.91$\pm$4.73  \\
        L-BN & 70.69$\pm$1.23 & 77.00$\pm$0.99 & 79.84$\pm$0.92 & 84.92$\pm$0.67 & 86.74$\pm$1.24 & 81.22$\pm$0.65 & 85.78$\pm$0.58 & 9.53$\pm$3.43 & 11.04$\pm$3.71  \\ 
        FIN & 73.03$\pm$1.45 & \textbf{78.04}$\pm$0.63 & \textbf{81.00}$\pm$0.38 & 86.12$\pm$0.33 & 88.02$\pm$0.49 & \textbf{82.15}$\pm$0.26 & 86.26$\pm$0.81 & 17.26$\pm$1.95 & 15.22$\pm$0.74  \\ 
		\bottomrule	
	\end{tabular}}
\end{table*}

\begin{table*}[!t]
	\centering
	\caption{\label{tbl:rnflt_gender}
        Performance on the test set of RNFLT maps with \textbf{gender} identities. Each experiment is run with 5 random seeds.
	    % Performance on the cross-sectional data for the glaucoma detection task. The baseline method is the supervised model trained with labeled samples only, while pseudo-sup stands for the proposed \textit{pseudo supervisor} method.
	}
 % \vspace{-0.3cm}
	\adjustbox{width=2\columnwidth}{
	\begin{tabular}{L{15ex} C{12ex} C{12ex} C{12ex} C{12ex} C{12ex} C{12ex} C{12ex} C{12ex}}
		\toprule
         & \textbf{Overall} & \textbf{Overall} & \textbf{Overall} & \textbf{Overall} & \textbf{Male} & \textbf{Female}  & \textbf{Overall} & \textbf{Overall} \\
		\textbf{Method} & \textbf{ES-Acc$\uparrow$} & \textbf{Acc$\uparrow$} & \textbf{ES-AUC$\uparrow$} & \textbf{AUC$\uparrow$} & \textbf{AUC$\uparrow$} & \textbf{AUC$\uparrow$} & \textbf{DPD$\downarrow$} & \textbf{DEOdds$\downarrow$} \\
		\cmidrule(lr){1-1} \cmidrule(lr){2-2} \cmidrule(lr){3-3} \cmidrule(lr){4-4} \cmidrule(lr){5-7} \cmidrule(lr){8-8} \cmidrule(lr){9-9}
        Adv~\cite{beutel2017data} & 75.39$\pm$1.54 & 77.18$\pm$1.45 & 83.87$\pm$1.45 & 84.62$\pm$1.16 & 84.31$\pm$1.36 & 85.22$\pm$1.01 & 6.15$\pm$0.38 & 7.92$\pm$0.49 \\
        SimCLR~\cite{chen2020improved} & 75.49$\pm$2.59 & 76.11$\pm$2.21 & 85.42$\pm$0.48 & 85.97$\pm$0.13 & 85.73$\pm$0.31 & 86.39$\pm$0.21 & 7.33$\pm$1.72 & 7.62$\pm$1.49 \\
        SupCon~\cite{khosla2020supervised} & 74.52$\pm$1.28 & 76.07$\pm$0.75 & 85.31$\pm$0.34 & 85.69$\pm$0.48 & 85.88$\pm$0.55 & 85.71$\pm$0.32 & 6.19$\pm$0.76 & 7.41$\pm$1.05 \\ \midrule
        FSCL~\cite{park2022fair} & 76.01$\pm$0.29 & 77.36$\pm$0.92 & 85.15$\pm$1.13 & 85.79$\pm$0.34& 86.09$\pm$0.16 & 85.51$\pm$0.98 & 6.21$\pm$0.68 & 7.63$\pm$0.49 \\
        FSCL+~\cite{park2022fair} & 75.93$\pm$1.18 & 76.93$\pm$1.87 & 85.37$\pm$0.38 & 85.92$\pm$0.58 & 85.98$\pm$0.79 & 85.83$\pm$0.40 & 6.13$\pm$0.85 & 7.07$\pm$1.41\\
        FSCL+ w/ FIN & 76.82$\pm$0.55 & 77.24$\pm$1.38 & 85.31$\pm$0.79 & \textbf{86.09}$\pm$0.61 & 85.63$\pm$0.24 & \textbf{86.54}$\pm$0.47 & \textbf{2.17}$\pm$0.81 & \textbf{1.85}$\pm$0.92
  \\
        \midrule
        No Norm & 76.67$\pm$1.18 & 77.80$\pm$0.32 & 83.83$\pm$0.49 & 85.05$\pm$0.19 & 84.87$\pm$0.87 & 85.40$\pm$0.77 & 4.79$\pm$1.82 & 5.79$\pm$1.83  \\
        BN \cite{Ioffe_ICML_2015} & 76.76$\pm$1.46 & 77.42$\pm$1.16 & 84.21$\pm$0.64 & 84.66$\pm$0.29 & 84.65$\pm$0.45 & 84.82$\pm$0.56 & 5.23$\pm$1.72 & 5.91$\pm$2.19  \\
        L-BN & 76.88$\pm$1.47 & 77.82$\pm$0.93 & 84.54$\pm$0.67 & 85.25$\pm$0.44 & 84.92$\pm$0.56 & 85.71$\pm$0.51 & 5.42$\pm$0.68 & 5.81$\pm$1.41 \\ 
        % Baseline+FIN & 77.16$\pm$1.09 & 77.80$\pm$0.74 & 85.47$\pm$0.35 & 85.67$\pm$0.37 & 85.66$\pm$0.38 & 85.75$\pm$0.49 & 3.47$\pm$2.27 & 4.11$\pm$2.39  \\ 
        FIN & \textbf{77.61}$\pm$0.67 & \textbf{78.44}$\pm$1.00 & \textbf{85.61}$\pm$0.25 & 86.04$\pm$0.23 & \textbf{86.17}$\pm$0.48 & 85.98$\pm$0.25 & 4.43$\pm$1.60 & 5.43$\pm$2.18 \\
		\bottomrule	
	\end{tabular}}
\end{table*}

\subsection{Equity-Scaled Metrics}
% IOU/(1+STD)
As discussed in Section \ref{sec:related}, conventional fairness metrics such as DDP and DEOdds fail to capture the equity-efficiency trade-off. For example, DDP=0 indicates perfect equity, but provides no information on performance metrics such as accuracy or AUC for each identity group. This limitation makes fairness metrics unsuitable for safety-critical medical applications. In such settings, it is essential to balance fairness and efficiency to ensure that the model is not only fair but also accurate and effective for all identity groups.

Hence, we propose an equity scaling measurement that takes both efficiency and equity into account for evaluation. The proposed measurement can easily and generally adapt to any conventional metrics, such as Accuracy, AUC, etc. Let $\mathcal{M} \in \{\text{Accuracy}, \text{AUC}, \ldots\}$ be a metric. Without loss of generality, conventional metrics like Accuracy and AUC take a set of triplets $(z', a, y)$ as input to produce the metric score, \ie, $\mathcal{M}(\{(z', y)\})$. These metrics ignore the identity information on the samples. When taking identity information into account, we can measure the performance of each identity group separately. Then, the performance discrepancy $\Delta$ between the whole set and identity groups can be obtained as follows
\begin{equation}
    \begin{split}
    \Delta = \sum_{A\in \mathcal{A}} &|\mathcal{M}(\{(z', y) \}) -\mathcal{M}(\{(z', a, y)|a=A \})|
\end{split}
\label{eqn:perf_delta}
\end{equation}
According to Equation (\ref{eqn:perf_delta}), $\Delta$ tends to 0 only if all the identity groups achieve similar performance. With $\Delta$, the equity-scaled metric $\text{ES-}\mathcal{M}$ is defined below
\begin{equation}
    \begin{split}
    \text{ES-}\mathcal{M} = \frac{\mathcal{M}(\{(z', y)\})}{1+\Delta}
\end{split}
\label{eqn:es_metric}
\end{equation}

There is a clear connection between $\text{ES-}\mathcal{M}$ and $\mathcal{M}$, \ie, $\text{ES-}\mathcal{M} \le \mathcal{M}$. Specifically, we have 
\begin{equation}
    \begin{split}
    \lim\limits_{\Delta\to 0} \text{ES-}\mathcal{M} = \lim\limits_{\Delta\to 0} \frac{\mathcal{M}(\{(z', y)\})}{1+\Delta} = \mathcal{M}
\end{split}
% \label{eqn:es_def}
\end{equation}
When $\Delta$ increases, $\text{ES-}\mathcal{M}$ would decrease. Conversely, $\text{ES-}\mathcal{M}$ would increase when $\Delta$ decreases. The advantage of the proposed $\text{ES-}\mathcal{M}$ is that it can be interpreted in the same way as $\mathcal{M}$. For example, suppose $\mathcal{M}=\text{Accuracy}$, a higher score of ES-Accuracy suggests that the predictor is not only more accurate but is also more equitable at the same time.

\REVISION{
The proposed equity-scaled metric is constrained within the bounds of the metrics AUC and Acc. Specifically, we express the ES-AUC as follows
\begin{equation}
    \begin{split}
    \text{ES-AUC} = \frac{\text{Overall AUC}}{1+\sum_{i}^{N}|\text{Overall AUC} - i \text{-th Group AUC}|},
\end{split}
\end{equation}
where $N$ is the number of groups. We know $0 \le |\text{Overall AUC} - i \text{-th Group AUC}| \le 1$. We can substitute this inequality back into the definition to establish the bounds:
\begin{equation}
    \begin{split}
    \frac{\text{Overall AUC}}{1+N} \le \text{ES-AUC} \le \text{Overall AUC} \le 1,
\end{split}
\end{equation}
Similar analysis can be applied to ES-Acc as well.
}

\section{Experiment \& Analysis}

\subsection{Set-Up}

\noindent\textbf{Dataset}. As we introduce in Section \ref{section:data}, 2100 RNFLT maps or OCT B-scans images are used for training, and 900 RNFLT maps or OCT Bscans images are used for evaluation. Also, we extensively evaluate the proposed method on RNFLT maps and OCT Bscans images. Apart from the labels indicating glaucoma/non-glaucoma, the samples include two types of social identities: race (\ie, Asia, Black, and White) and gender (\ie, male and female). Please note that all our checkpoints in the paper are selected based on the validation set. The validation set comprises 100 Asian samples, 100 Black samples, and 100 White samples. The overall experimental results on the validation set align with those obtained from the test set. We have released the validation set along with the training set and the test set. 

\begin{table*}[!t]
	\centering
	\caption{\label{tbl:oct_race}
	    Performance on the test set of OCT B-scans images with race identities. All experiments used the same random seed.
     % The baseline method is the supervised model trained with labeled samples only, while pseudo-sup stands for the proposed \textit{pseudo supervisor} method.
	}
 % \vspace{-0.3cm}
	\adjustbox{width=2\columnwidth}{
	\begin{tabular}{L{15ex} C{12ex} C{12ex} C{12ex} C{12ex} C{12ex} C{12ex} C{12ex} C{12ex} C{12ex}}
		\toprule
         & \textbf{Overall} & \textbf{Overall} & \textbf{Overall} & \textbf{Overall} & \textbf{Asian} & \textbf{Black} & \textbf{White} & \textbf{Overall} & \textbf{Overall} \\
		\textbf{Method} & \textbf{ES-Acc$\uparrow$} & \textbf{Acc$\uparrow$} & \textbf{ES-AUC$\uparrow$} & \textbf{AUC$\uparrow$} & \textbf{AUC$\uparrow$} & \textbf{AUC$\uparrow$} & \textbf{AUC$\uparrow$} & \textbf{DPD$\downarrow$} & \textbf{DEOdds$\downarrow$} \\
		\cmidrule(lr){1-1} \cmidrule(lr){2-2} \cmidrule(lr){3-3} \cmidrule(lr){4-4} \cmidrule(lr){5-8} \cmidrule(lr){9-9} \cmidrule(lr){10-10}
        No Norm & 70.04 & 76.89 & 79.02 & 86.95 & 89.29 & 81.66 & \textbf{89.36} & \textbf{10.67} & 7.90 \\
        BN \cite{Ioffe_ICML_2015} & 60.30 & 66.33 & 77.44 & 84.86 & 86.45 & 79.73 & 87.72 & 11.33 & \textbf{6.63}  \\
        L-BN & 72.05 & 77.33 & 77.45 & 85.24 & 87.37 & 79.56 & 87.48 & 14.67 & 17.35  \\ 
        FIN & \textbf{74.42} & \textbf{78.22} & \textbf{81.19} & \textbf{87.14} & \textbf{89.58} & \textbf{82.70} & 87.60 & 17.00 & 11.50  \\ 
		\bottomrule	
	\end{tabular}}
\end{table*}

\begin{table*}[!t]
	\centering
	\caption{\label{tbl:oct_gender}
	    Performance on the test set of OCT B-scans images with gender identities.  All experiments are with the same seed.
     % The baseline method is the supervised model trained with labeled samples only, while pseudo-sup stands for the proposed \textit{pseudo supervisor} method.
	}
 % \vspace{-0.3cm}
	\adjustbox{width=2\columnwidth}{
	\begin{tabular}{L{15ex} C{12ex} C{12ex} C{12ex} C{12ex} C{12ex} C{12ex} C{12ex} C{12ex}}
		\toprule
         & \textbf{Overall} & \textbf{Overall} & \textbf{Overall} & \textbf{Overall} & \textbf{Male} & \textbf{Female} & \textbf{Overall} & \textbf{Overall} \\
		\textbf{Method} & \textbf{ES-Acc$\uparrow$} & \textbf{Acc$\uparrow$} & \textbf{ES-AUC$\uparrow$} & \textbf{AUC$\uparrow$} & \textbf{AUC$\uparrow$} & \textbf{AUC$\uparrow$} & \textbf{DPD$\downarrow$} & \textbf{DEOdds$\downarrow$} \\
		\cmidrule(lr){1-1} \cmidrule(lr){2-2} \cmidrule(lr){3-3} \cmidrule(lr){4-4} \cmidrule(lr){5-7} \cmidrule(lr){8-8} \cmidrule(lr){9-9}
        No Norm & 75.42 & 76.56 & 84.36 & 86.12 & 85.26 & 87.35 & \textbf{3.40} & 4.30 \\
        BN \cite{Ioffe_ICML_2015} & 73.13 & 73.89 & 82.77 & 83.79 & 83.22 & 84.45 & 4.83 & 4.91 \\
        % L-BN & 75.87 & 76.89 & 84.68 & 86.53 & 85.51 & 87.69 & 6.57 & 7.03  \\ 
        L-BN & 74.86 & 77.44 & 85.22 & 86.68 & 85.93 & 87.65 & 5.17 & 7.67 \\
        FIN &  \textbf{79.27} & \textbf{79.44} & \textbf{85.61} & \textbf{87.11} & \textbf{86.30} & \textbf{88.05} & 3.94 & \textbf{3.52}  \\ 
		\bottomrule	
	\end{tabular}}
\end{table*}

\noindent\textbf{Method}. EfficientNet-B1 \cite{Tan_ICML_2019} is used as the baseline model to handle RNFLT maps, while 3D ResNet-18 \cite{Yang_JBHI_2021} is used as the baseline model to handle 3D OCT B-scans images. Except for the baseline model with no normalization prior to the final linear layer (No Norm), we plug batch normalization (BN), learnable batch normalization (L-BN), and the proposed FIN ($m=0.3$) to the baseline model before the final linear layer for comprehensive comparison purpose.
% Specifically, learnable batch normalization is similar to batch normalization, which applies a z-normalization to the features. Different from batch normalization, $\mu$, and $\sigma$ are not empirically computed with the batch of the features but are learned from the features along the learning process. 
\REVISION{
Specifically, given a mini-batch $\mathcal{B}$, the batch normalization is defined as $\widehat{z}=\frac{z-\mu_\mathcal{B}}{\sigma_\mathcal{B}}$, where $\mu_\mathcal{B}$ and $\sigma_\mathcal{B}$ are the mean and standard deviation computed from the mini-batch.
In contrast, the learnable batch normalization has additional learnable parameters for the mean and standard deviation instead of computing them from each mini-batch. In particular, learnable batch normalization is defined as:
$\widehat{z}=\frac{z-\mu}{\sigma}$
where $\mu$ and $\sigma$ are learnable parameters that are optimized along with the model weights to minimize the loss, capturing batch-level statistics. Unlike standard batch normalization, which computes $\mu_\mathcal{B}$ and $\sigma_\mathcal{B}$ from each batch, the learnable statistics in learnable batch normalization estimate the distribution of the entire dataset. This allows greater flexibility for the model to learn the appropriate parameters for normalizing mini-batches.
}

For comparative methods, we select adversarial training~\cite{beutel2017data}, and four contrastive losses including SimCLR~\cite{chen2020improved} and SupCon~\cite{khosla2020supervised} to investigate how unfair contrastive learning performs on our RNFLT data, and fair supervised contrastive loss~\cite{wang2022fairness} (FSCL) and its variant FSCL+ that combines contrastive loss and their attribute group-wise normalization loss~\cite{wang2022fairness}. For fair comparisons, we adopt the same EfficientNet (B1) as the backbone and use the default experimental setups and hyper-parameters from their official codebases. 

\noindent\textbf{Metric}. To fully understand the balance between efficiency and equity, we utilize multiple metrics, including Accuracy, AUC, DPD \cite{Agarwal_ICML_2018,Agarwal_ICML_2019}, and DEOdds \cite{Agarwal_ICML_2018}.
% \REVISION{Specifically, DPD is defined as the difference between the largest and the smallest group-level selection rate $\tau(A)=\mathbb{E}[h(x) \mid a=A]$, \ie, $\tau(A_{max}) - \tau(A_{min})$ where $h(x)$ is a predicted label, and $A_{max}=\argmax_{A}\tau(A), A_{min}=\argmin_{A}\tau(A)$, across all values of the sensitive attribute(s). The DPD of 0 means that all groups have the same selection rate. Equalized odds aim to examine if the machine learning model’s predictions are not only independent of sensitive group membership, but that groups have the same false positive rates $P[h(x)=1 \mid a=A, Y=0]$ and true positive rates $P[h(x)=1 \mid a=A, Y=1]$. DEOdds consists of the difference of false positive rates and the difference of true positive rates. The DEOdds of 0 means that all groups have the same true positive, true negative, false positive, and false negative rates.
% }
\REVISION{DPD measures the maximum difference in selection rates across groups defined by sensitive attributes. Specifically, DPD is defined as $\text{DPD} = \tau(A_{\max}) - \tau(A_{\min})$ where $\tau(A) = \mathbb{E}[h(x) \mid a = A]$ is the selection rate for group A, $A_{\max} = \argmax_{A} \tau(A)$ is the group with max selection rate, $A_{\min} = \argmin_{A} \tau(A)$ is the group with min selection rate, $h(x)$ is the predicted label, A DPD of 0 indicates all groups have equal selection rates. Equalized odds aims for quantify similarity between false positive rates ($\text{FPR}(A) = P[h(x)=1 \mid a = A, Y=0] $) and true positive rates ($\text{TPR}(A) =  P[h(x)=1 \mid a = A, Y=1]$) across groups. The difference in equalized odds is the maximum difference in FPR and TPR between groups. Equalized odds is achieved when the difference in odds is 0, meaning equal FPR and TPR for all groups.
}
Additionally, we introduce ES-Acc and ES-AUC to measure the impact of different identities on efficiency, specifically Accuracy and AUC. Since AUC is a crucial performance metric in medical applications, we also calculate identity-specific AUC values to assess unfairness in the data.

\noindent\textbf{Training Scheme}. We train models with various fairness and normalization methods using an AdamW optimizer \cite{Loshchilov_ICLR_2019} and an NVIDIA RTX A6000 Graphics Card. In our experiments with RNFLT maps, we use a learning rate of 5e-5, a weight decay of 0, and betas (0, 0.1). For OCT B-scans images, we use a learning rate of 1e-5, a weight decay of 0, and the same betas. We set the batch size to 6 for RNFLT maps and 2 for OCT B-scans images. The hyperparameters are determined by optimizing the performance of the baseline model. 

During contrastive baseline training, we follow SimCLR \cite{chen2020improved} and FSCL \cite{wang2022fairness} to apply data augmentation techniques to SimCLR~\cite{chen2020improved}, SupCon~\cite{khosla2020supervised}, FSCL~\cite{wang2022fairness}, and FSCL+~\cite{wang2022fairness} due to the requirements of using the contrastive learning loss. The contrastive pre-training stage is trained for 30 epochs with a batch size of 128 and Adam optimizer, and the fine-tuning stage takes 20 epochs. We set 5e-3 as the learning rate for pre-training.  For fine-tuning, we use the same setup as our proposed methods. 
On the other hand, we use a data augmentation technique on No Norm, BN, L-BN, and FIN. The model is trained with RNFLT maps for 10 epochs and OCT B-scans images for 30 epochs. The baseline model with No Norm, BN, L-BN, and FIN uses the cross-entropy loss as the objective function.

\begin{figure}[!t]
	\centering
	\subfloat[]{\includegraphics[width=0.23\textwidth]{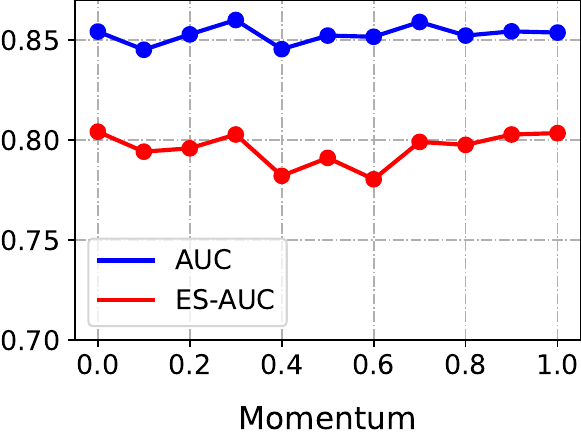}  \label{fig:momentum_auc_race}} \hfill
	\subfloat[]{\includegraphics[width=0.23\textwidth]{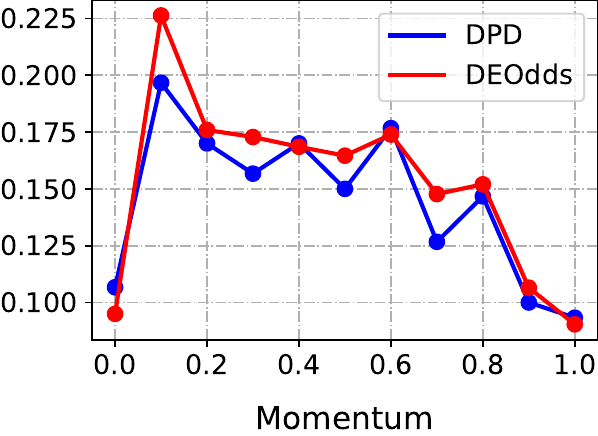}  \label{fig:momentum_dpd_race}} \\
        \subfloat[]{\includegraphics[width=0.23\textwidth]{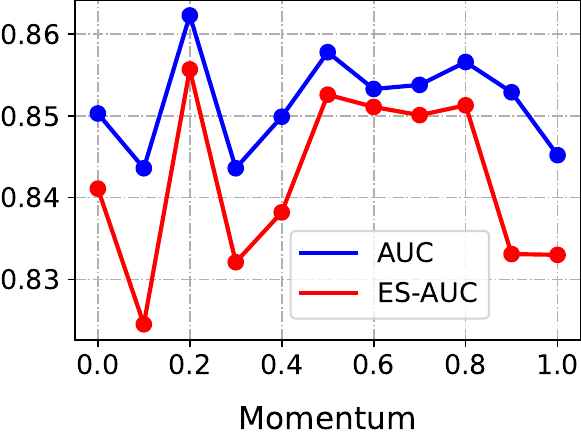} \label{fig:momentum_auc_gender}} \hfill
        \subfloat[]{\includegraphics[width=0.23\textwidth]{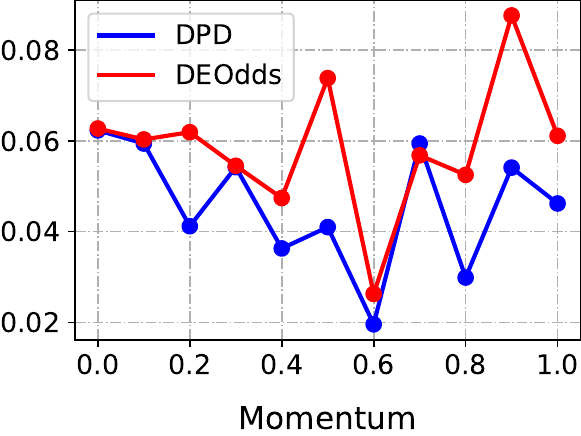}
        \label{fig:momentum_dpd_gender}}
	\caption{\label{fig:momentum}
    	Ablation study of momentum $m$ in Equation (\ref{eqn:momentum}) on RNFLT maps. $m$ ranges from 0 to 1 with step 0.1. (a) and (b) are based on the experiment with the race identities, while (c) and (d) are based on the experiment with the gender identities. 
    	}
\end{figure}

\subsection{Efficiency \& Fairness Evaluation}
Table \ref{tbl:rnflt_race} reports the experimental results on RNFLT maps with respect to racial identities, respectively. The results indicate that the AUC for Blacks is consistently lower than the AUC for Whites across all methods. Additionally, the AUC for Whites is generally lower than the AUC for Asians. These findings suggest that automated glaucoma detection with deep learning is particularly challenging in the Black group. Compared to the baseline model (No Norm), the proposed FIN improves the performance of glaucoma detection in the Black group by 2.69\% (p-value $=$ 0.002). Moreover, the FIN outperforms the other methods in terms of the metrics ES-Acc, Acc, ES-AUC, and AUC. Comparing the performance of FIN with BN/L-BN indicates that normalizing all features with the same identity leads to a better performance gain than normalizing all features in the same batch. This finding aligns with the hypothesis that features with the same identity are highly correlated.

In addition, the fairness metrics DPD and DEOdds exhibit contradictory patterns with Acc and AUC. For instance, FSCL+ achieves the lowest DPD and DEOdds, but its corresponding Acc and AUC are significantly lower than the ones yielded by the proposed FIN. In contrast, the proposed ES-Acc and ES-AUC not only align with Acc and AUC, but also reflect equity across racial identities.

In contrast to Table \ref{tbl:rnflt_race}, where the differences between the three racial identities are significant (p-value $=$ 0.002), Table \ref{tbl:rnflt_gender} demonstrates that the difference between male and female groups is relatively small. This leads to overall higher ES-Acc scores and ES-AUC scores than the ones in Table \ref{tbl:rnflt_race}. The finding is aligned with the definition of the proposed equity scaling measurement (\ie, Equation (\ref{eqn:es_metric})).

Similar to the results shown in Table \ref{tbl:rnflt_race}, the proposed FIN improves the performance of glaucoma detection over metrics such as Acc and AUC, even for DPD and DEOdds when compared to the baseline model. Moreover, ES-Acc and ES-AUC are generally consistent with Acc and AUC across all methods.

The role of data augmentation techniques or adversarial fairness loss in the 3D baseline model is unclear. As a result, Adv, SimCLR, SupCon, FSCL, and FSCL+ are not used to evaluate 3D OCT B-scans images. Table \ref{tbl:oct_race} presents the performance on OCT B-scans images w.r.t. race identities. It shows a similar pattern as Table \ref{tbl:rnflt_race}, where the Black group exhibits lower AUC than the other two groups. Once again, the proposed FIN improves the AUC for the Black group.

\begin{table}[!t]
	\centering
\caption{\label{tbl:val_rnflt_race}
	    Performance on the \textbf{validation} set of RNFLT maps with \textbf{race} identities.
	}
	\adjustbox{width=1\columnwidth}{
	\begin{tabular}{L{15ex} C{12ex} C{12ex} C{12ex} C{12ex} C{12ex} }
		\toprule
          & \textbf{Overall} & \textbf{Overall} & \textbf{Asian} & \textbf{Black} & \textbf{White}  \\
		\textbf{Method}  & \textbf{ES-AUC$\uparrow$} & \textbf{AUC$\uparrow$} & \textbf{AUC$\uparrow$} & \textbf{AUC$\uparrow$} & \textbf{AUC$\uparrow$}  \\ \midrule
        Adv  & 75.88 & 83.57 & 81.97 &  79.44 & 88.65 \\
        SimCLR& 78.90 & 84.97  & 85.52 & 80.78 & 87.15  \\
        SupCon  & 79.32  & 85.49  & 84.21 & 82.29  & 88.66  \\ \midrule
        FSCL   & 77.88  & 84.32 &84.74 & 78.82 & 87.26  \\
        FSCL+  &  77.48 &  83.94 & 83.96  & 79.23 & 86.67 \\
        FSCL+ w/ FIN &  80.14 & 86.55 & \textbf{86.81} & 81.58 & 89.32 \\ \midrule
        % No Norm & 70.37 & 76.00 &	78.63 & 84.82 & 87.22 & 80.65 & 86.11 & 15.00 & 13.12  \\
        No Norm  & 77.10 & 83.56 & 84.42 & 78.82 & 86.35 \\
        BN  & 79.97 & 83.43 & 83.89 & 81.44 & 85.31  \\
        L-BN  & 80.36 & 84.06 & 85.71 & 81.21 & 84.15  \\ 
        FIN & \textbf{80.77} & \textbf{86.93} & 86.12 & \textbf{82.98} & \textbf{89.80}  \\
		\bottomrule	
	\end{tabular}}
 % \vspace{-0.6cm}
\end{table}

\begin{table}[!t]
	\centering
 \caption{\label{tbl:rnflt_marital}
	    Performance on the \textbf{test} set of RNFLT maps with \textbf{marital status} identities.
     % All scores are in the form of percentages. 
     % The baseline method is the supervised model trained with labeled samples only, while pseudo-sup stands for the proposed \textit{pseudo supervisor} method.
	}
 
\adjustbox{width=1.\linewidth}{
	\begin{tabular}{L{15ex}  C{10ex} C{10ex} C{10ex} C{10ex} C{10ex} C{10ex} C{10ex} }
		\toprule
         & \textbf{Overall} & \textbf{Overall} & \textbf{Married} & \textbf{Single} & \textbf{Divorced} & \textbf{Widowed} & \textbf{Leg-Sep}  \\
		\textbf{Method} & \textbf{ES-AUC$\uparrow$} & \textbf{AUC$\uparrow$} & \textbf{AUC$\uparrow$} & \textbf{AUC$\uparrow$} & \textbf{AUC$\uparrow$} & \textbf{AUC$\uparrow$} & \textbf{AUC$\uparrow$}  \\ \midrule
		% \cmidrule(lr){1-1} \cmidrule(lr){2-2} \cmidrule(lr){3-3} \cmidrule(lr){4-4} \cmidrule(lr){5-8} 
        Adv & 58.97 & 85.52 & 86.84 &  83.22 & 85.81 & 89.57 & 48.12  \\
        SimCLR & 59.32  & 85.83  & 87.17 & 83.35  & 83.39 & \textbf{94.19} & 55.66 \\
        SupCon & 70.29 & 85.42 & 85.12 & 85.52 & 83.21 & 90.49 & 70.22   \\ \midrule
        FSCL  & 68.32 & 85.07  & 84.33 & \textbf{89.32} & 83.98  & 90.22 & 61.39 \\
        FSCL+  & 54.24 & 85.42 & 87.21 & 82.73 & 83.53  & 88.24 & 37.20   \\ 
        FSCL+ w/ FIN  & 71.40 & \textbf{86.97} & \textbf{87.70} & 85.74 & 85.39 & 85.29 & 70.37   \\ 
        % 71.40 & 86.97 & 87.70 & 85.74 & 85.39 & 85.29 & 70.37
        \midrule
        
        No Norm  & 62.89 & 84.44 & 85.99 & 82.85 & 78.60 & 88.24 & 62.96 \\
        BN  & 62.04 & 84.94 & 86.14 & 82.36 & 85.39 & 88.24 & 55.56 \\
        L-BN  & 74.32 & 84.66 & 84.82 & 83.57 & 85.60 & 89.50 & 77.78  \\ 
        FIN & \textbf{79.03} & 86.22 & 86.23 & 84.76 & \textbf{87.45} & 91.60 & \textbf{85.19}\\ 
		\bottomrule	
	\end{tabular}}
\end{table}

\begin{table}[!t]
	\centering
\caption{\label{tbl:rnflt_ethnicity}
	    Performance on the test set of RNFLT maps with \textbf{ethnicity} identities. 
	}
	\adjustbox{width=1\linewidth}{
	\begin{tabular}{L{15ex} C{12ex} C{12ex} C{12ex} C{12ex}}
		\toprule
         & \textbf{Overall} & \textbf{Overall} & \textbf{Non-Hisp} & \textbf{Hispanic}  \\
		\textbf{Method}  & \textbf{ES-AUC$\uparrow$} & \textbf{AUC$\uparrow$} & \textbf{AUC$\uparrow$} & \textbf{AUC$\uparrow$}  \\ \midrule
        Adv  & 83.69 & 85.38 &  85.23 & 83.37   \\
        SimCLR &  80.32 & 86.09 & 77.13 & 84.29  \\
        SupCon   & 79.28  & 86.21 & 77.15  & 88.95   \\ \midrule
        FSCL   & 81.65 & 84.25 & 84.21  & 87.59   \\
        FSCL+   & 83.72 & 85.53 & 85.22 & 87.49  \\
        FSCL+ w/ FIN  & 83.53 & 86.00 & 85.94 & 88.89  \\
        % 0.714	0.8697	0.877	0.8574	0.8539	0.8529	0.7037
        \midrule
        No Norm  & 76.10 & 84.71 & 84.52 & \textbf{95.83}  \\
        BN  & 77.43 & 84.96 & 84.72 & 94.44  \\
        L-BN  & 79.02 & 84.67 & 84.52 & 91.67 \\ 
        FIN  & \textbf{85.02} & \textbf{86.76} & \textbf{86.76} & 84.72  \\ 
		\bottomrule	
	\end{tabular}}
\end{table}

Table \ref{tbl:oct_gender} displays a similar pattern to Table \ref{tbl:rnflt_gender}, where the performances w.r.t. the male and female groups are comparable. The proposed FIN can enhance the performances w.r.t. both groups. Moreover, the proposed ES-Acc and ES-AUC metrics align more closely with Acc and AUC. This is because the difference between the two groups is relatively small. 

In Tables \ref{tbl:rnflt_marital} and \ref{tbl:rnflt_ethnicity}, we present the fairness results associated with marital status and ethnicity, respectively, utilizing AUC as the metric of interest. The proposed FIN stands out by achieving the highest ES-AUC score for both categories. This trend is consistent with our previous observations in other tables, where we found that fair adversarial and contrastive losses play a crucial role in enhancing marginal fairness across diverse groups.

\REVISION{
Regarding marital status, disparities are most evident, particularly between the widowed and legally separated groups, showing the largest AUC discrepancies, ranging from 40\% to 50\%.
}
However, our FIN strategy effectively raises the AUC performance of the legally separated group to an impressive 85.19\%, thereby reducing the AUC disparity with the widowed group to a minimum of 6.4\%.

% In the case of marital status, the unfairness is more pronounced than in other identity categories, with the largest AUC discrepancies, ranging from 40\% to 50\%, noted between the widowed and legally separated groups. However, our FIN strategy effectively raises the AUC performance of the legally separated group to an impressive 85.19\%, thereby reducing the AUC disparity with the widowed group to a minimum of 6.4\%.

Regarding ethnicity, our FIN proves its value by enhancing the performance of the non-Hispanic group while preserving a comparable AUC for the Hispanic group. This outcome contributes to a substantial augmentation in the ES-AUC, thus further illustrating the efficacy of our proposed approach. 

\REVISION{
% In addition to apply FIN to the training process, we also validate if the proposed FIN can work in the pre-trainiing process, \eg, with FSCL \cite{wang2022fairness}. We denote it as FSCL+ w/ FIN. As shown in Table \ref{tbl:rnflt_race}, FSCL+ w/ FIN improves the performance on all the metrics, compared to FSCL+. Similarly, we observe consistent performance improvement on multiple metrics in Table \ref{tbl:rnflt_gender}, \ref{tbl:rnflt_marital}, and \ref{tbl:rnflt_ethnicity}. This verifies the effectiveness of the proposed FIN for minimizing demographic disparities.
In addition, we explore the application of the proposed FIN not only during the training but also in the pre-training phase, exemplified by its integration with FSCL+ \cite{wang2022fairness}, denoted as FSCL+ w/ FIN. As depicted in Table \ref{tbl:rnflt_race}, FSCL+ w/ FIN exhibits enhanced performance across all metrics compared to FSCL+. Similar improvement on multiple metrics can be consistently observed in Table \ref{tbl:rnflt_gender}, \ref{tbl:rnflt_marital}, and \ref{tbl:rnflt_ethnicity}, demonstrating the efficacy of the proposed FIN in mitigating demographic disparities throughout various demographic attributes. These results demonstrate that the potential of our FIN module can be used with other existing model fairness-promoting techniques to collectively improve model equity.
}

We next present the fairness performance metrics for various racial groups, utilizing the \textbf{validation set}. It's important to note that all the checkpoints mentioned in this paper are chosen based on the results derived from this validation set, which consists of 100 Asian samples, 100 Black samples, and 100 White samples. From the findings presented in Table \ref{tbl:val_rnflt_race}, we can observe that the overall performance patterns gleaned from the validation set correspond closely with those acquired from the test set. 

Last but not least, using 3D OCT B-scans images leads to better performance on the metric AUC than using RNFLT maps with the baseline model and the baseline model with the proposed FIN.

% Point: BN L-BN show that identity-wise vs batch-wise 

% Point 1: DPD, DEOdds cannot reflect the trade-off

% Point 2: The proposed equity scaling method can work with Acc and AUC.

% Point 3: EyeFair reflects unfairness in medical applications. The data with race identities are more unfair than the ones with gender identities.  

% Point 4: OCT B-scans images lead to better performance than RNFLT maps.

% Point 5: ES-AUC with Race identities is lower than the one with Gender identities. 

\subsection{Analysis}

\noindent\textbf{Ablation Study}.
As introduced in Section \ref{sec:method}, the proposed FIN has three hyperparameters, \ie, $\mu_{A}$, $\sigma_{A}$, and $m$ where $A\in \mathcal{A}$. Note that $\mu_{A}$ are $\sigma_{A}$ are randomly initialized according to a unit Gaussian distribution. We provide the ablation study of $m$, which ranges from 0 to 1 with step 0.1. To comprehensively understand its effects on both efficiency and fairness, we present the plots of the momentum vs. AUC/ES-AUC, and the plots of the momentum vs. DPD/DEOdds.

As shown in Figure \ref{fig:momentum_auc_race}, $m=0.3$ achieves the best performance on AUC and ES-AUC with the race identities. Figure \ref{fig:momentum_auc_gender} shows that $m=0.2$ achieves the best performance on AUC and ES-AUC with the gender identities. Therefore, without loss of generality, we set $m$ to 0.3 for the experiments.  
On the other hand, we can see that the curve of ES-AUC is consistent with the one of AUC. In contrast, DPD and DEOdds exhibit different pattern from AUC and ES-AUC. The curves of DPD and DEOdds display significant variability as the value of $m$ is changed.

\begin{figure}[!t]
	\centering
	\subfloat[Baseline]{\includegraphics[width=0.23\textwidth]{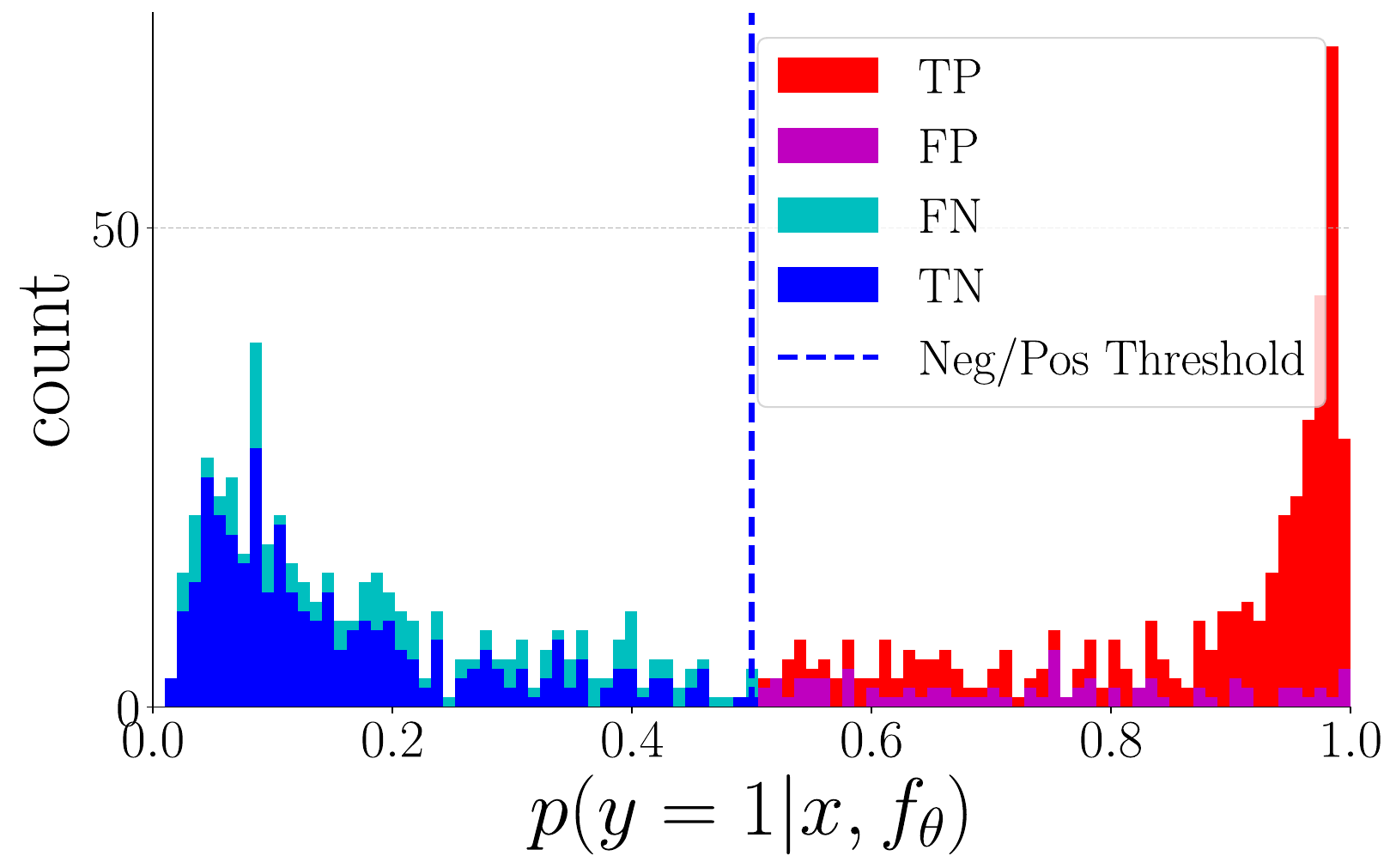}  \label{fig:hist_base}} \hfill
	\subfloat[Baseline with BN]{\includegraphics[width=0.23\textwidth]{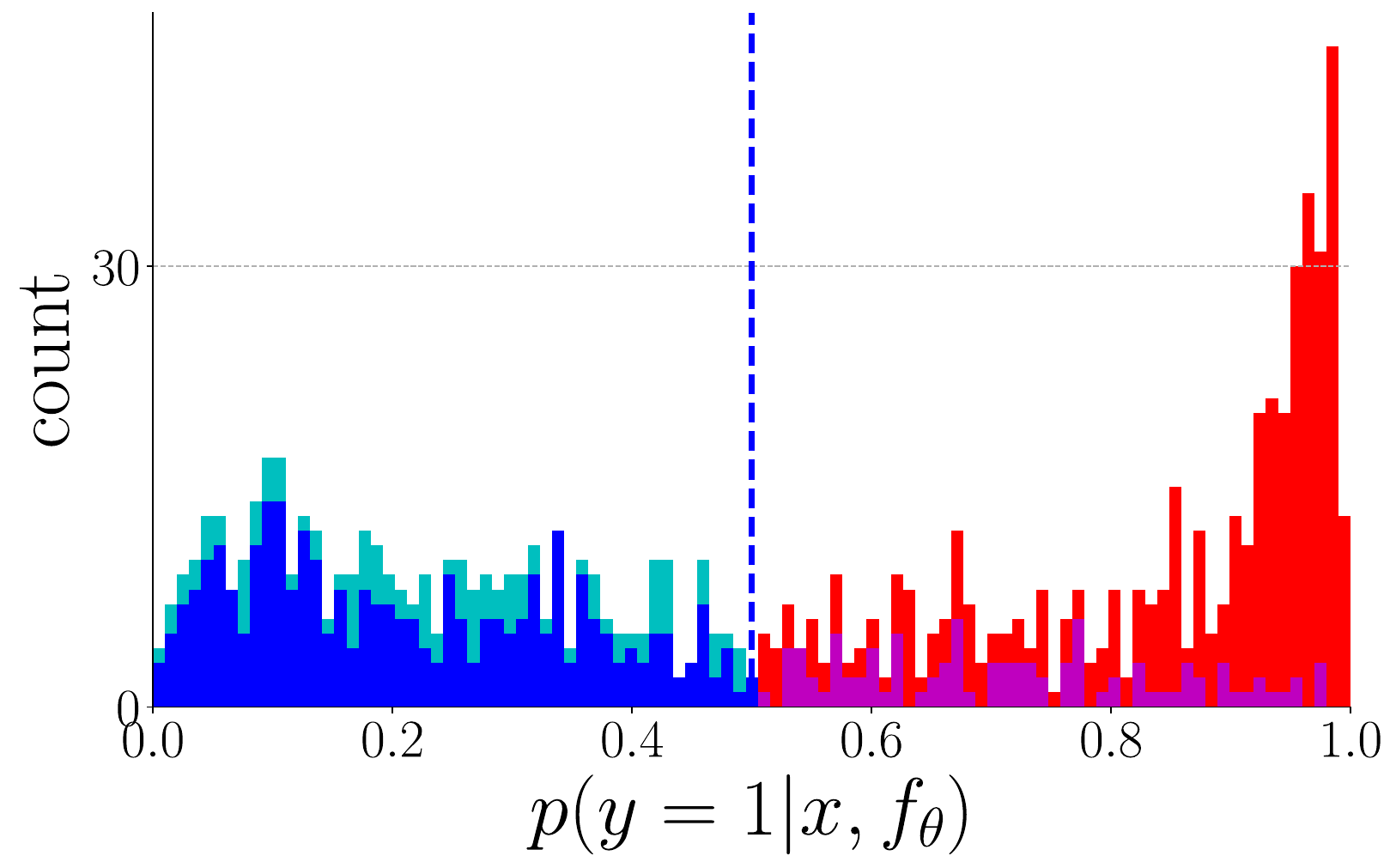}  \label{fig:hist_base_bn}} \\
        \subfloat[Baseline with L-BN]{\includegraphics[width=0.23\textwidth]{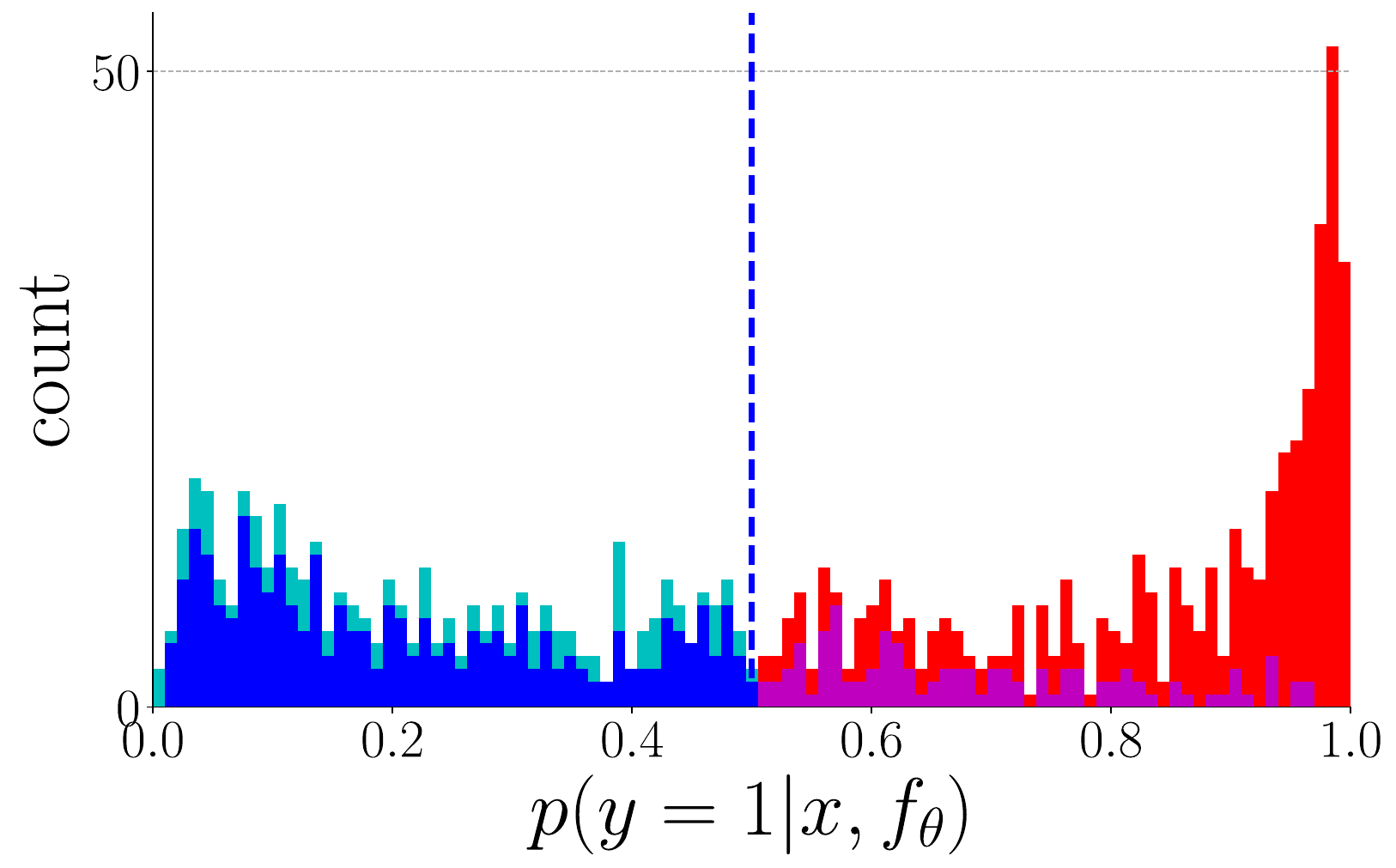} \label{fig:hist_base_lbn}} \hfill
        \subfloat[Baseline with FIN]{\includegraphics[width=0.23\textwidth]{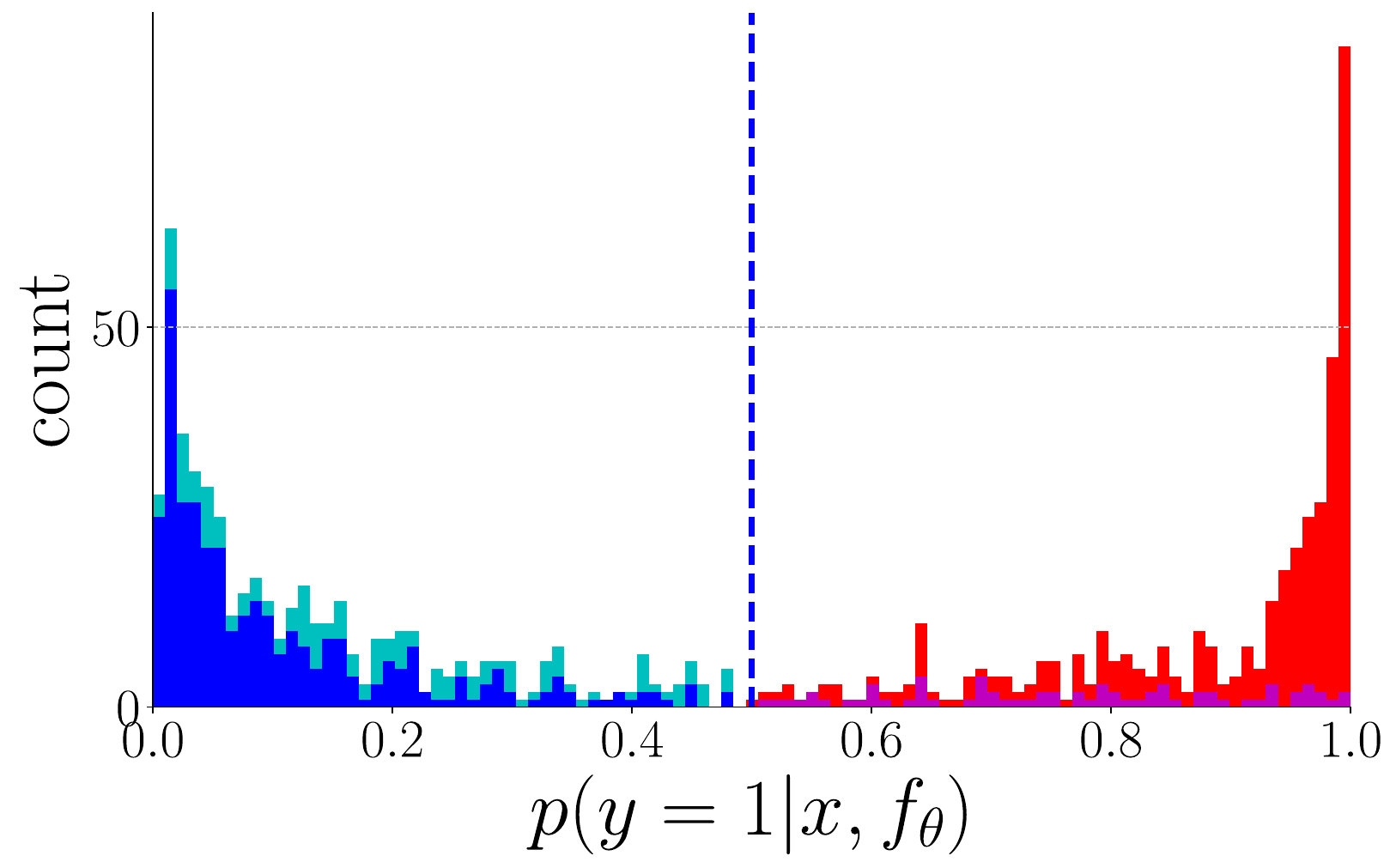}
        \label{fig:hist_base_fin}}
	\caption{\label{fig:hist}
    	Histograms of predictions yielded by the baseline, baseline with batch normalization, baseline with learnable batch normalization, and baseline with fair identity normalization on RNFLT maps. TP, FP, TN, and FN stand for true positive, false positive, true negative, and false negative, respectively.
    	}
\end{figure}

\noindent\textbf{Effects of FIN}.
% To explore the effects of the proposed FIN on the learning process, we visualize the TPs, FPs, TNs, and FNs of the predictions yielded by the baseline, baseline with BN, baseline with L-BN, and baseline with FIN. As shown in Figure \ref{fig:hist}, the proposed FIN pushes TPs and TNs towards the right and the left, respectively. This is because the proposed FIN enhances the features according to its identity-specific characteristics. 
To examine the impact of the proposed FIN on the learning process, we visualize the true positives (TPs), false positives (FPs), true negatives (TNs), and false negatives (FNs) of the predictions generated by the baseline model, the baseline model with BN, the baseline model with L-BN, and the baseline model with FIN. Figure \ref{fig:hist} illustrates that the proposed FIN shifts TPs and TNs to the right and left, respectively. This is because the proposed FIN enhances the features based on their identity-specific characteristics.

\begin{figure}[!t]
  \centering
    \includegraphics[width=0.25\textwidth]{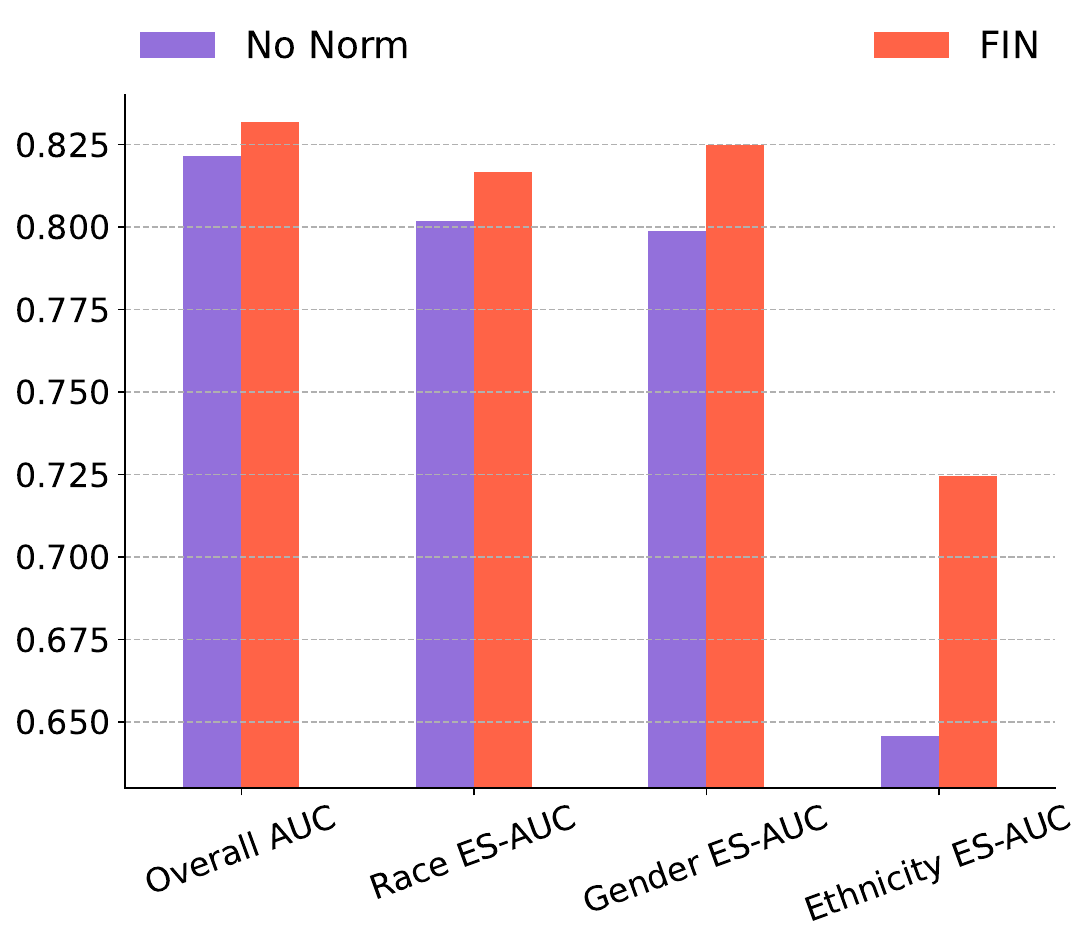}
    \vspace{-0.2cm}
  \caption{Performance of the proposed FIN with the attribute spherical equivalent (SE), compared with the baseline No Norm. Emmetropia is defined as SE within the range of -0.5 to +0.5 D, hyperopia as greater than +0.5 D, and myopia as less than -0.5 D.} 
   \vspace{-0.7cm}
  \label{fig:myopia}
\end{figure}

\noindent\textbf{Influence of Myopia on Fairness}.
\REVISION{
High myopia, in particular, has been identified as a significant factor contributing to an elevated risk of primary open-angle glaucoma \cite{Chen_IJO_2012}. Adhering to the criteria set by Lavanya et al., \cite{Lavanya_IOVS_2010}, emmetropia is characterized by a spherical equivalent (SE) within the range of -0.5 to +0.5 D, hyperopia as exceeding +0.5 D, and myopia as falling below -0.5 D. With SE information available for 1,159 patients, they are categorized into these three groups. Figure \ref{fig:myopia} illustrates the proposed FIN's performance compared to the baseline No Norm. The normalization of features by the SER attribute results in enhanced performance, as indicated by improvements in overall AUC and ES-AUC w.r.t. race, gender, and ethnicity. This aligns with established findings highlighting the association between myopia and glaucoma. Furthermore, our results suggest that we may use SE information as a surrogate variable to implicitly represent gender, race, and ethnicity for fairness learning when this demographic information is missing.
}
\section{Conclusion}

 While minority groups experience more health issues, there are currently no dedicated medical datasets with imaging data available for fairness learning, though deep learning relies heavily on imaging data. This paper presents Harvard-GF, a retinal nerve disease dataset for detecting glaucoma enabling fairness learning, which is the leading cause of irreversible blindness worldwide disproportionately affecting Blacks. We proposed a fair identity normalization (FIN) approach to equalize feature importance between different identity groups to improve model fairness with superior performance to various SOTA models. We designed an equity-scaled performance metric to evaluate model performance penalized by model fairness.

\section{Acknowledgments}
This work was supported by NIH R00 EY028631, R21 EY035298, P30 EY003790, Alcon Young Investigator Grant, and Research to Prevent Blindness International Research Collaborators Award.

{\small
\bibliographystyle{IEEEtran}
\bibliography{egbib}
}

\end{document}